\documentclass[3p]{elsarticle}

\usepackage{amsmath,amssymb}
\usepackage{graphicx}
\usepackage{booktabs}
\usepackage{multirow}
\usepackage{algorithm}
\usepackage{algorithmic}
\usepackage{hyperref}
\usepackage{float}
\usepackage{placeins}
\usepackage[font=small,labelfont=bf,labelsep=period]{caption}

\graphicspath{{}}

\newcommand{\SemKVBlockSpace}{\par\vspace{0.55\baselineskip}}
\BeforeBeginEnvironment{figure}{\SemKVBlockSpace}
\AfterEndEnvironment{figure}{\SemKVBlockSpace}
\BeforeBeginEnvironment{table}{\SemKVBlockSpace}
\AfterEndEnvironment{table}{\SemKVBlockSpace}
\BeforeBeginEnvironment{algorithm}{\SemKVBlockSpace}
\AfterEndEnvironment{algorithm}{\SemKVBlockSpace}
\BeforeBeginEnvironment{equation}{\SemKVBlockSpace}
\AfterEndEnvironment{equation}{\SemKVBlockSpace}
\BeforeBeginEnvironment{equation*}{\SemKVBlockSpace}
\AfterEndEnvironment{equation*}{\SemKVBlockSpace}
\BeforeBeginEnvironment{align}{\SemKVBlockSpace}
\AfterEndEnvironment{align}{\SemKVBlockSpace}
\BeforeBeginEnvironment{align*}{\SemKVBlockSpace}
\AfterEndEnvironment{align*}{\SemKVBlockSpace}
\BeforeBeginEnvironment{gather}{\SemKVBlockSpace}
\AfterEndEnvironment{gather}{\SemKVBlockSpace}
\BeforeBeginEnvironment{gather*}{\SemKVBlockSpace}
\AfterEndEnvironment{gather*}{\SemKVBlockSpace}
\BeforeBeginEnvironment{multline}{\SemKVBlockSpace}
\AfterEndEnvironment{multline}{\SemKVBlockSpace}
\BeforeBeginEnvironment{multline*}{\SemKVBlockSpace}
\AfterEndEnvironment{multline*}{\SemKVBlockSpace}

\makeatletter\def\ps@pprintTitle{\let\@oddhead\@empty\let\@evenhead\@empty\def\@oddfoot{\footnotesize\itshape Preprint. Under review.\hfill\today}\let\@evenfoot\@oddfoot}\makeatother
\begin{document}

\begin{frontmatter}

\title{SemKV: Semantic Mixed-Precision KV Cache Quantization Guided by the Quality Cliff for Long-Context LLM Inference}

\author[inst1]{Daeha Lee\corref{cor1}}
\ead{bigsum@etri.re.kr}
\author[inst1]{Do-Hyung Kim}
\author[inst1]{Jae-Hong Kim}
\cortext[cor1]{Corresponding author.}
\affiliation[inst1]{organization={Electronics and Telecommunications Research Institute (ETRI)},
            city={Daejeon},
            country={Republic of Korea}}

\begin{abstract}
The key--value (KV) cache is a primary memory bottleneck in long-context large language model inference, since its size grows linearly with context length. Existing KV-cache compression spans uniform quantization, token deletion, and importance-aware mixed precision, but it remains unclear how to choose precision levels without crossing a sharp quality boundary --- and deletion additionally risks removing evidence that long-context reasoning depends on. We introduce SemKV, an all-token-preserving mixed-precision KV-cache quantization framework guided by an empirically measured \emph{quality cliff}.

Sweeping uniform quantization over a fractional-level grid under a prespecified multi-seed statistical protocol, we find that Llama-3.1-8B-Instruct with an affine scalar quantizer is statistically indistinguishable from FP16 KV down to 2.322 code bits/value and degrades sharply at 2.0 bits. SemKV measures this boundary offline, then assigns two adjacent above-cliff precisions (2.585/2.322 bits) to tokens by a model-internal importance ranking, realizing an average of 2.39 code bits/value (2.65 effective bits/value after packing and metadata) --- a measured $6.0\times$ storage reduction with no statistically detectable quality difference from FP16 KV on our sampled LongBench evaluation ($n{=}900$, three seeds). Interior probes show that the tested partial-protection mixtures do not recover full-KV-level quality when they cross the cliff, and a block-wise deferred allocation extends the mechanism to generated tokens and multi-turn dialogue. Preserving all tokens at low precision also substantially outperforms FP16 token pruning even though the pruning baseline uses a $1.5\times$ larger memory footprint, and the cliff, the interpolation property, and the operating point transfer to Mistral-7B-Instruct-v0.3, where informative selection remains necessary for safe mixing.

Finally, the measured cliff depends on the model--quantizer pair under a specified evaluation protocol: replacing the affine base with a distortion-optimized quantizer (TurboQuant-MSE) lowers the collapse boundary in every protocol we test, and SemKV converts the extra headroom directly into compression --- interpolating at 2.0/1.585 bits inside the newly opened grid gap, it is statistically indistinguishable from full KV at an effective 2.025 bits/value, raising the no-detectable-loss operating point from $6.0\times$ to $7.9\times$. A small above-cliff floor persists in full-cache multi-turn quantization under both bases (its magnitude $2.5\times$ smaller under TurboQuant). In the tested multi-turn comparisons, SemKV remains statistically indistinguishable from the upper grid point while using fewer bits. These results motivate a general recipe: measure the cliff for the target deployment setting, then interpolate above it.
\end{abstract}

\begin{keyword}
KV cache compression \sep Long-context inference \sep Large language models \sep Mixed precision quantization \sep Semantic importance
\end{keyword}

\end{frontmatter}

\FloatBarrier

\section{Introduction}

Large language models (LLMs) increasingly operate on contexts of tens of thousands of tokens for document QA, code understanding, retrieval-augmented generation, and multi-turn agents~[31--36, 44--46]. In Transformer decoding, the key--value (KV) cache grows linearly with context length, and its memory footprint governs the feasible batch size and context budget of practical serving systems~[37--38, 47]. Uniform KV quantization~[1--2, 9], eviction and pruning~[13--24, 55], and efficient attention~[25--29] have all been proposed, but uniform quantization is blind to token-level differences, and deletion-based methods risk removing evidence tokens.

This paper starts from an empirical observation rather than an architectural one. When uniform KV quantization is swept over a \emph{fractional-bit} grid (e.g., $\log_2 5 \approx 2.322$, $\log_2 6 \approx 2.585$ bits realized by integer quantization levels), quality does not degrade gradually: on our LongBench protocol with Llama-3.1-8B-Instruct, all uniform settings from 3.0 down to 2.322 bits are statistically indistinguishable from full-precision KV, and quality collapses only in the interval $(2.0, 2.322]$ --- a \emph{quality cliff} that co-occurs with a sharp increase in runaway generation. Under the same affine base, the same collapse location reappears in single-turn QA, in generation-token quantization, and in multi-turn dialogue (where a small above-cliff residual floor additionally exists), and even on a different backbone (Mistral-7B-Instruct-v0.3).

The cliff reframes what mixed precision is for. In the tested setting, mixtures crossing the cliff fail to recover full-KV performance, whereas allocation has little measurable effect when both precision levels remain above it. The useful regime is the sparse fractional-bit grid just above the cliff, where uniform quantization can only sit \emph{on} grid points, while mixed precision can \emph{interpolate between} them. SemKV therefore preserves every token, scores each token with a model-internal semantic importance indicator, and assigns the higher of two above-cliff grid precisions to the top-ranked fraction, reaching average precisions (e.g., 2.39 bits between 2.322 and 2.585) that uniform quantization cannot realize, at quality statistically indistinguishable from full KV.

The main contributions of this paper are as follows.
\begin{itemize}
\item \textbf{Quality-cliff map on a fractional-bit grid.} We chart uniform KV quantization on a fractional-bit grid under a prespecified multi-seed statistical protocol and locate a quality cliff in $(2.0, 2.322]$ bits --- whose catastrophic transition is reproduced across prefill, generation, and multi-turn protocols under this base (multi-turn additionally shows a small above-cliff floor; Sec.~4.6) --- for Llama-3.1-8B, co-occurring with runaway generation; on the primary single-turn LongBench protocol the region $[2.322, 3.0]$ is statistically indistinguishable from full KV.
\item \textbf{Grid interpolation by all-token-preserving mixed precision.} SemKV keeps all tokens and interpolates between above-cliff grid points via importance-ranked two-level allocation, reaching otherwise unreachable average precisions with no statistically detectable deficit from full KV --- a measured $6.0\times$ KV-storage reduction versus FP16 at the 2.39-bit operating point, metadata included. Ablations over eight model-internal indicators (hidden-state, KV-norm, logit, and attention families) show the effect is structural in the flat regime: indicator choice does not change the outcome there.
\item \textbf{Generation-token and multi-turn extension.} A block-exact deferred-quantization mechanism extends token-wise mixed precision to generation-time tokens; under the affine base, generation quality degrades sharply at 2.0 bits whereas 2.322-bit generation remains statistically indistinguishable from FP16, and the extension holds across multi-turn dialogue.
\item \textbf{Deletion versus low-precision preservation.} Under matched (in fact adverse) memory budgets, FP16 pruning collapses (0.065--0.149) while SemKV at 2.39 bits shows no statistically detectable difference from full KV (0.438); random pruning is statistically indistinguishable from indicator-based pruning and exhibits the same qualitative collapse, implicating token deletion as the dominant source of failure in this controlled comparison.
\item \textbf{Stress and transfer.} A widened-gap stress test whose low-bit side lies below the cliff produces a clear ranking among the tested model-internal indicators, showing that indicator quality becomes important when most tokens receive a below-cliff precision, and the cliff and interpolation property transfer to Mistral-7B-Instruct-v0.3, where SemKV remains statistically indistinguishable from full KV when informative indicators are used; replacing the base quantizer with TurboQuant-MSE shows the cliff is quantizer-dependent, and that the downward shift is protocol-wide: the collapse boundary moves one grid step down to $(1.585, 2.0]$ in prefill and multi-turn, and to $(1.0, 1.585]$ on the generation side, while SemKV's token-axis allocation composes unchanged, matches TurboQuant's channel-axis 2.5-bit recipe at a smaller effective footprint, and occupies the affine-inaccessible 1.9-code-bit gap at performance statistically indistinguishable from full KV ($7.9\times$).
\end{itemize}
\section{Related Work}

\subsection{KV Cache Quantization}

In long-context LLM inference, the KV cache grows in proportion to the input length and becomes a major GPU memory bottleneck when long contexts and large batch sizes are used. To address this issue, recent KV-cache quantization methods have been actively developed. KVQuant~[1] targets sub-4-bit KV cache quantization and combines per-channel quantization for the key cache, pre-RoPE key quantization, non-uniform datatypes, and outlier-aware dense-and-sparse quantization to achieve low performance degradation even at the 3-bit level. KIVI~[2] analyzes the distributional characteristics of the KV cache and proposes an asymmetric 2-bit KV cache quantization method that quantizes the key cache per channel and the value cache per token. TurboQuant~[9] takes a quantizer-design view: it applies a random rotation so that coordinates approach a known distribution, quantizes each coordinate with a distribution-matched optimal (Lloyd--Max) scalar quantizer, and optionally corrects inner-product bias with a 1-bit residual sketch, achieving near-optimal distortion without calibration data; its fractional operating points (e.g., 2.5 bits) are realized by \emph{channel-wise} mixed precision that quantizes outlier channels at a higher bit-width. TurboQuant is therefore orthogonal to SemKV along two axes: it improves the per-vector quantizer while allocating precision across \emph{channels}, whereas SemKV allocates precision across \emph{tokens} on a fractional-level grid; the two compose, and Sec.~4.9 evaluates SemKV with TurboQuant as its base quantizer.

General LLM quantization studies are also related to SemKV. SmoothQuant~[3], GPTQ~[4], AWQ~[5], LLM.int8()~[6], ZeroQuant~[7], QServe~[8], QuaRot~[10], Atom~[11], block reconstruction-based post-training quantization~[56], and post-training quantization~[57-59] aim to improve LLM inference efficiency through weight or activation quantization. However, these studies mainly focus on tensor-, channel-, or weight-level quantization. In contrast, SemKV differs in that it assigns KV cache precision differently based on token-level semantic importance.

A line of work closer to SemKV applies \emph{mixed-precision} KV quantization. MiKV~[71] retains would-be-evicted KV pairs at low precision and important pairs at high precision, with importance imported from external eviction policies such as H2O; QAQ~[72] adapts bit allocation to token importance and sensitivity; SKVQ~[73] combines clipped dynamic quantization with a high-precision recency window; ZipCache~[74] identifies salient tokens via normalized attention scores; GEAR~[75] compresses most entries to ultra-low precision with low-rank residual correction; WKVQuant~[76] and RotateKV~[77] preserve recent or sink tokens at high precision; IntactKV~[78] keeps outlier-token caches intact. Recent analysis also studies the token-count versus precision trade-off directly~[79]. SemKV differs from these methods in three respects. First, SemKV treats token importance as a pluggable model-internal signal and systematically evaluates eight indicators across four families (hidden-state, KV, logit, and attention), rather than committing to a single eviction policy or fixed positional heuristic. Second, SemKV explicitly maps the uniform-quantization quality cliff on a fractional-bit grid and positions mixed precision as \emph{grid interpolation strictly above the cliff}, which explains \emph{when} allocation matters (near or below the cliff) and when it does not (the flat regime). Third, the same ranking-based mechanism is extended block-exactly to generation-time tokens and multi-turn inference under a prespecified multi-seed statistical protocol.

\subsection{KV Cache Compression and Eviction}

Another direction for reducing KV cache memory is eviction or compression, which keeps only a subset of KV cache tokens and removes the rest. Based on the observation that heavy-hitter tokens are repeatedly and importantly referenced in attention, H2O~[13] proposes a KV cache eviction policy that retains both recent tokens and heavy-hitter tokens. SnapKV~[14] observes attention patterns within an observation window before generation and selects important KV positions on which each attention head focuses. PyramidKV~[15] exploits the phenomenon of pyramidal information funneling, in which layer-wise information flow is broadly distributed in lower layers and concentrated on core tokens in upper layers. KVzip~[81] scores KV pairs by their necessity for reconstructing the original context, enabling query-agnostic eviction whose compressed cache remains reusable across subsequent queries. FastGen~[17], ChunkKV~[18], InfLLM~[19], RetrievalAttention~[20], context compression~[66], and semantic-aware cache compression~[67] also address KV cache or attention compression for long-context inference.

Although these studies are effective in reducing the size of the KV cache, they commonly rely on selecting or removing a subset of KV tokens. In contrast, SemKV does not remove tokens. SemKV preserves all tokens and adjusts only their precision according to importance.

\subsection{Token Pruning and Sparse Inference}

Token pruning and sparse inference studies aim to reduce computation and memory by decreasing the number of input tokens or intermediate tokens. LazyLLM~[21], DynamicViT~[22], Token Merging~[23], TokenLearner~[24], and Adaptive Token Pruning~[55] perform dynamic token selection or merging based on token importance. Sparse Transformer~[25], Longformer~[26], and Big Bird~[27] make the attention pattern itself sparse to enable processing of long sequences.

However, pruning or sparse selection can remove tokens themselves or restrict their accessibility. Since SemKV preserves all tokens and adjusts only their precision, it is an all-token-preserving compression method that is fundamentally different from pruning.

\subsection{Efficient Attention and Long-Context Inference}

FlashAttention~[28] and FlashAttention-2~[29] greatly improve the efficiency of attention computation through IO-aware exact attention. RoFormer~[30] improves the positional representation of Transformers through rotary position embeddings. StreamingLLM~[16] uses the attention sink phenomenon to enable stable inference even for long streaming sequences. ALiBi~[65] addresses the difficulty of extrapolating Transformers to contexts longer than those seen during training.

These studies improve attention computation or positional generalization, whereas SemKV is a complementary approach in that it compresses the KV cache memory itself using semantic mixed precision during the decoding stage.

\subsection{Long-Context Evaluation}

Benchmarks such as LongBench~[40], LongBench v2~[42], and InfiniteBench~[43] have been proposed to evaluate long-context LLMs. Lost in the Middle~[41] showed that LLMs do not always use long input contexts uniformly and that performance may degrade when answer information is located in the middle of the context. These studies show that evidence position, retrieval difficulty, and distractor composition are important in long-context evaluation. Based on this line of work, this paper evaluates SemKV on QA/retrieval tasks, a recency stress test, pruning comparison, and a mixed LongBench boundary setting.

\subsection{LLM Serving, Memory Optimization, and Mixed Precision}

To address memory bottlenecks in LLM inference and training, FlexGen~[12], vDNN~[49], ZeRO~[50], ZeRO-Infinity~[51], Megatron-LM~[52], studies on large-scale training systems~[53-54], GShard~[61], Switch Transformer~[62], PaLM~[63], and Chinchilla~[64] have been proposed. In addition, mixed-precision training~[48] and adaptive mixed-precision inference~[68] demonstrate the importance of precision allocation. SemKV extends this mixed-precision perspective to the token dimension of the KV cache.

\subsection{Position of SemKV}

Taken together, existing methods for reducing KV cache memory can be broadly divided into three categories. First, uniform KV quantization quantizes all tokens with the same bit-width~[1-2, 9]. Second, pruning- or eviction-based methods remove tokens considered less important~[13-14, 21, 55]. Third, some methods retain a subset of tokens based on specific positions or attention patterns, such as attention sinks, recent tokens, and heavy-hitter tokens~[16, 41]. SemKV occupies a different position from these approaches. SemKV does not remove tokens; it preserves all tokens. At the same time, it does not quantize all tokens uniformly, but assigns higher precision to tokens with higher semantic importance. Table~1 summarizes this positioning.

\FloatBarrier

\begin{table}[H]
\centering
\caption{Positioning against representative KV compression families. Examples: uniform quantization --- KIVI, KVQuant; eviction --- H2O, SnapKV; importance-aware quantization --- MiKV, QAQ, ZipCache. SemKV is distinguished by the combination of all-token preservation, token-wise fractional-grid precision mapped from a measured cliff, and generation-time support. Entries summarize common characteristics of representative methods and do not imply that every method in a family has all listed properties.}
\small
\setlength{\tabcolsep}{4pt}
\resizebox{\textwidth}{!}{%
\begin{tabular}{lccccc}
\toprule
Method family & \shortstack[c]{Deletes\\tokens} & \shortstack[c]{Per-token\\bits} & Indicator & \shortstack[c]{Cliff-mapped\\frac.\ grid} & Gen.-time \\
\midrule
Uniform quantization & No & No & --- & No & Partial \\
Eviction / pruning & Yes & No & Attn. & No & Yes \\
Importance-aware quant. & Varies & Yes & Attn. & No & Partial \\
SKVQ & No & Window & Recency & No & Yes \\
\textbf{SemKV (this work)} & \textbf{No} & \textbf{Yes} & \shortstack[c]{\textbf{Model-internal}\\\textbf{(8 tested)}} & \textbf{Yes} & \shortstack[c]{\textbf{Yes}\\\textbf{(block-exact)}} \\
\bottomrule
\end{tabular}}%
\end{table}
\section{SemKV Method}

This section describes the proposed SemKV method. The central objective of SemKV is to reduce KV cache memory usage in long-context LLM inference while preserving as much semantic information as possible for long-context question answering.

The overall architecture of SemKV is shown in Fig.~1. The input prompt passes through the Transformer prefill stage to generate a full-precision KV cache. SemKV then computes token-level semantic importance and separates high-precision and low-precision tokens according to the importance ranking. Finally, it constructs a mixed-precision KV cache using the token-wise bit allocation results and uses it during decoding. Through this structure, SemKV reduces KV cache memory without token pruning while retaining all token positions and prioritizing tokens with high model-internal importance scores.

\subsection{Overall Architecture}

\begin{figure}[H]
\centering
\includegraphics[width=\textwidth]{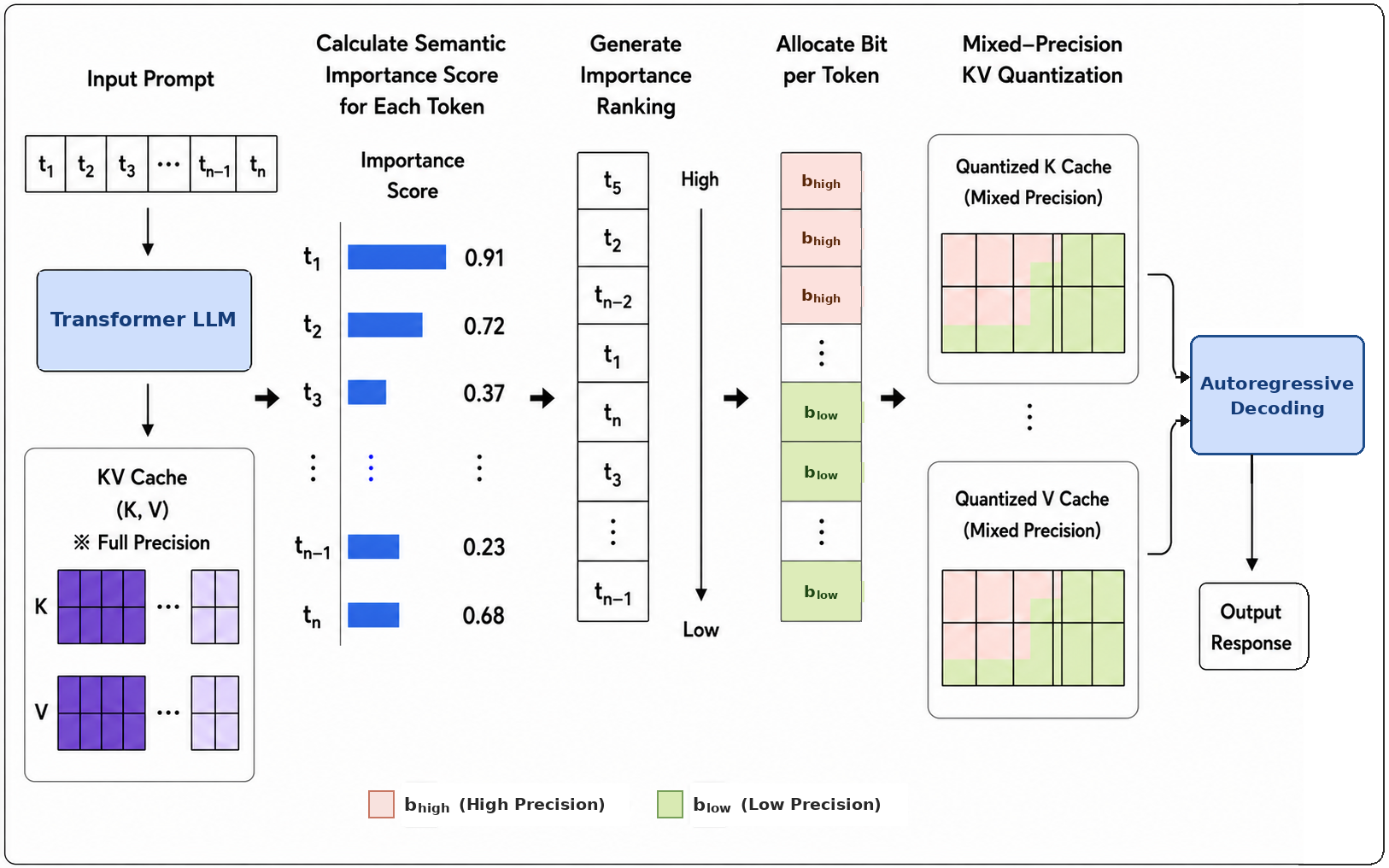}
\caption{Overall SemKV architecture. SemKV first generates the full-precision KV cache, estimates token-level semantic importance, ranks tokens, allocates higher precision to important tokens, and performs mixed-precision KV quantization before decoding. The two precision levels shown inside the panel ($b_{\mathrm{high}}$/$b_{\mathrm{low}}$) are set in our experiments to fractional-level codes such as 2.585/2.322 bits (Sec.~4.1).}
\label{fig:overall}
\end{figure}
\FloatBarrier
\subsection{Problem Formulation}

Let the input sequence length be $T$, the number of Transformer layers be $L$, the number of key--value heads be $H_{\mathrm{KV}}$ (for grouped-query-attention models such as Llama-3.1 this is smaller than the attention-head count), the dimension of each head be $D$, and the KV cache precision be $b$ bits. Then, for a single sequence and ignoring quantization metadata, the KV cache storage in bits can be approximated as follows.
\begin{equation}
M_{\mathrm{KV}} = 2 \cdot L \cdot H_{\mathrm{KV}} \cdot T \cdot D \cdot b .
\end{equation}

SemKV keeps the sequence length $T$ unchanged while assigning a different token-wise bit-width $b_i$.
\begin{equation}
T_{\mathrm{SemKV}} = T,\quad
b_i \in \{b_{\mathrm{low}}, b_{\mathrm{high}}\}.
\end{equation}

As illustrated in Fig.~2, SemKV estimates semantic importance from model-internal signals --- by default, hidden-state dynamics from the final Transformer layers. Rather than relying on token position or recency heuristics, SemKV analyzes how token representations evolve across late layers. Tokens with larger representation changes are hypothesized to contribute more strongly to contextual processing, and we use this quantity as an importance proxy in long-context QA.

\subsection{Semantic Importance Estimation}

\begin{figure}[H]
\centering
\includegraphics[width=\textwidth]{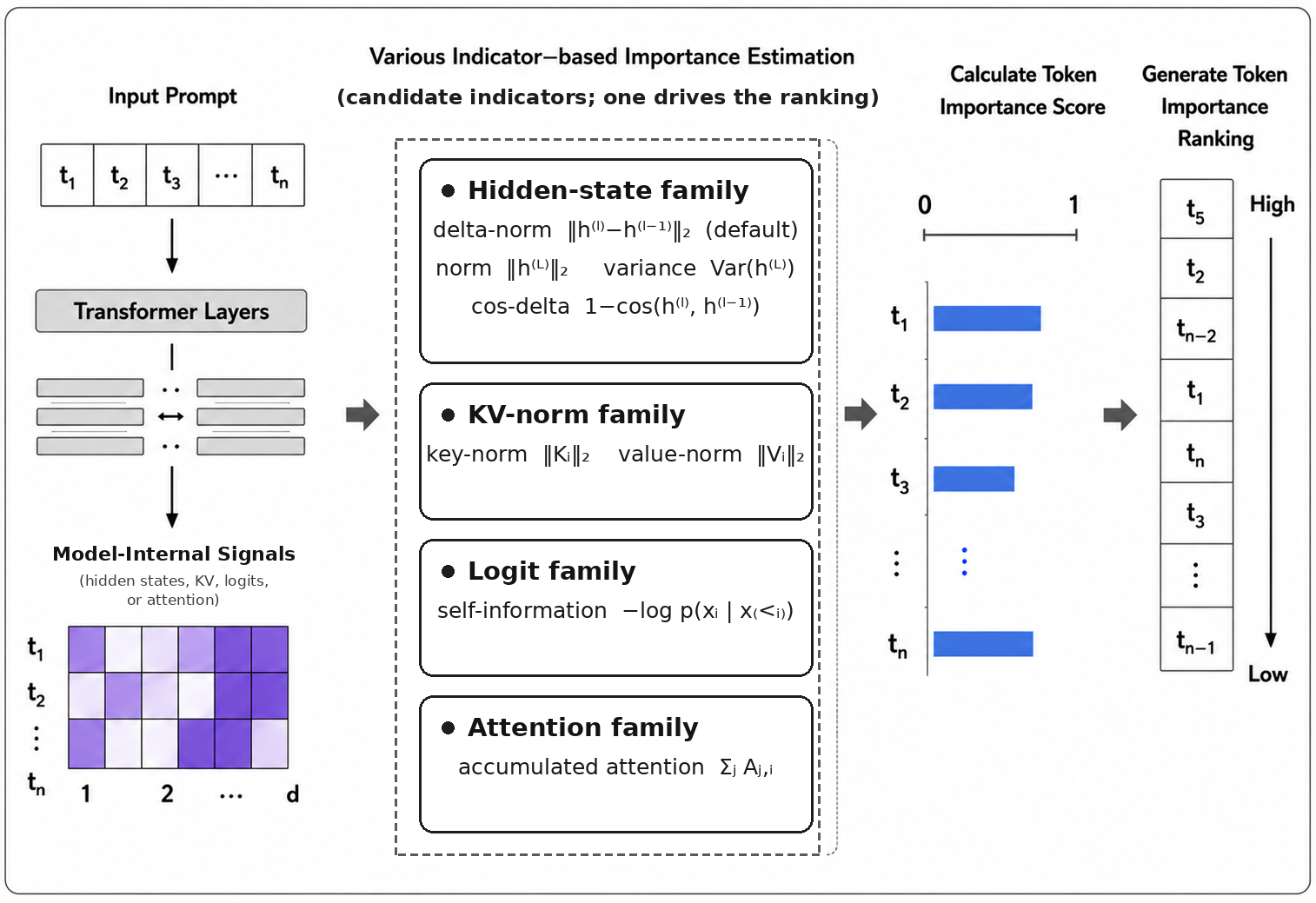}
\caption{Semantic importance scoring pipeline. SemKV computes a token-importance score from one selected model-internal indicator and ranks tokens within each prompt; because allocation depends only on the ranking, score normalization is not required (the 0--1 score scale in the panel is illustrative).}
\label{fig:scoring}
\end{figure}
\FloatBarrier
For an input token $i$, let the hidden representation at layer $l$ be $h_i^{(l)}$. When the last $m$ layers are used, the delta-based semantic importance score of SemKV can be defined as follows.
\begin{equation}
s_i =
\frac{1}{m}
\sum_{l=L-m+1}^{L}
\left\| h_i^{(l)} - h_i^{(l-1)} \right\|_2 .
\end{equation}

\noindent Equation~(3) defines the default \emph{delta-norm} indicator. SemKV treats the indicator as a pluggable module and, throughout the ablations, draws on four families of model-internal signals, all obtained from tensors the model already computes. The \emph{hidden-state} family reads the last-$m$-layer representations: besides delta-norm, it includes the final-layer magnitude (\emph{norm}, $\|h_i^{(L)}\|_2$), the coordinate-wise \emph{variance} of $h_i^{(L)}$, and the angular counterpart of Eq.~(3) (\emph{cosine-delta}, $1-\cos(h_i^{(l)}, h_i^{(l-1)})$ averaged over the last $m$ layers), which captures direction change even when magnitudes are stable. The \emph{KV-norm} family scores each token by the magnitude of the very tensors being compressed (\emph{key-norm} $\|K_i\|_2$ and \emph{value-norm} $\|V_i\|_2$), a proxy for how strongly a token can contribute to attention logits and outputs. The \emph{logit} family uses \emph{self-information}, $-\log p_\theta(x_i \mid x_{<i})$: tokens that the model finds surprising tend to introduce new information. Finally, the \emph{attention} family uses accumulated attention received by token $i$ ($\sum_{j \ge i} A_{j,i}$), the signal closest to eviction heuristics such as H2O; unlike the other families it requires materializing attention weights, which drives the cost analysis of Sec.~4.5. A \emph{random} selector serves as the uninformative baseline in ablations. Scores enter the method only through their within-prompt ranking, so no normalization is required, and the allocation procedure of Sec.~3.4 remains identical across indicators. We restrict the average to the \emph{final} $m$ layers under the heuristic that, as depth increases, Transformer hidden states integrate progressively broader contextual information about each token than earlier layers; we treat Eq.~(3) as a practical proxy for a token's semantic contribution to the context, the quantity the allocation should protect.

\subsection{Token-Wise Mixed-Precision Allocation}

\noindent\textbf{Operating rule: map the cliff first.} The bit-precision pair $(b_{high}, b_{low})$ consumed by the allocation of Sec.~3.4 is not a free hyperparameter: it is determined by an offline \emph{cliff-mapping} procedure performed once for each fixed deployment setting (model, base quantizer, evaluation protocol, and context-budget composition), illustrated in Fig.~3. The procedure sweeps uniform quantization over the integer-level grid, compares each grid point against the full-precision cache with paired statistics on a representative evaluation set, examines adjacent grid-point differences, and locates the \emph{quality cliff} --- a sharp, seed-consistent increase in degradation between two adjacent grid points, below which output quality collapses catastrophically rather than degrading gracefully (Sec.~4.2). $(b_{high}, b_{low})$ are then chosen as grid points above the measured cliff; because no uniform grid point exists between adjacent levels, the token-mixed allocation of Fig.~4 provides a practical mechanism for realizing average bit-widths inside that gap. The mapping is input-independent --- it characterizes the target deployment setting, not individual prompts --- and need not be repeated for individual prompts, but it should be revalidated when the model, base quantizer, evaluation protocol, or context-budget composition changes materially; Sec.~4.9 shows that replacing the quantizer moves the boundary downward in all three protocols studied in this paper --- prefill compression (Sec.~4.2), generation-time quantization (Sec.~4.4), and multi-turn dialogue (Sec.~4.6) --- with the same operating rule carrying over; under the affine base the bracketed interval coincides across protocols, while under TurboQuant the generation-side boundary sits one further grid step lower.

\begin{figure}[H]
\centering
\includegraphics[width=\textwidth]{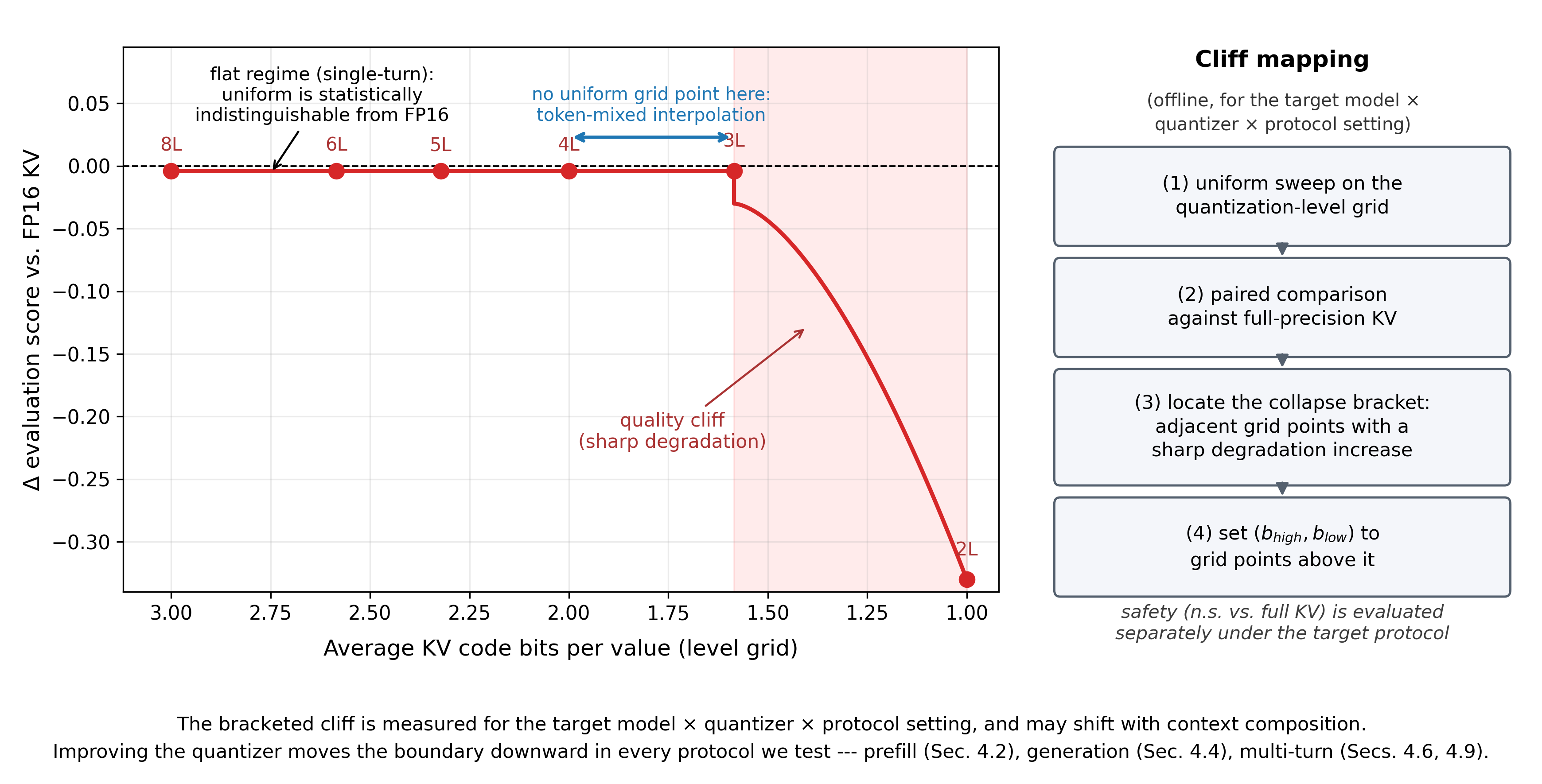}
\caption{The quality-cliff concept and the offline mapping procedure. Uniform quantization on the level grid remains in a relatively flat regime until degradation increases sharply between two adjacent grid points; small residual deficits may exist above this collapse bracket, particularly in full-cache multi-turn evaluation (Sec.~4.6). In the single-turn setting illustrated here, the above-cliff grid points are statistically indistinguishable from full precision. The boundary is an empirically measured property of the model, base quantizer, and target evaluation setting; improving the quantizer shifts it downward in every protocol we test (the generation-side boundary may sit lower than the prefill and multi-turn boundaries). The curve is schematic; only the grid points are empirically evaluated. SemKV sets $(b_{high}, b_{low})$ to grid points above the measured cliff and interpolates between them.}
\end{figure}

Fig.~4 illustrates the adaptive mixed-precision allocation process in SemKV. After computing semantic importance scores, SemKV ranks tokens and allocates higher precision to the most important subset of tokens. The remaining tokens are quantized using lower precision. This strategy allows SemKV to significantly reduce average KV precision while retaining all token positions and prioritizing tokens with high importance scores.

\begin{figure}[H]
\centering
\includegraphics[width=\textwidth]{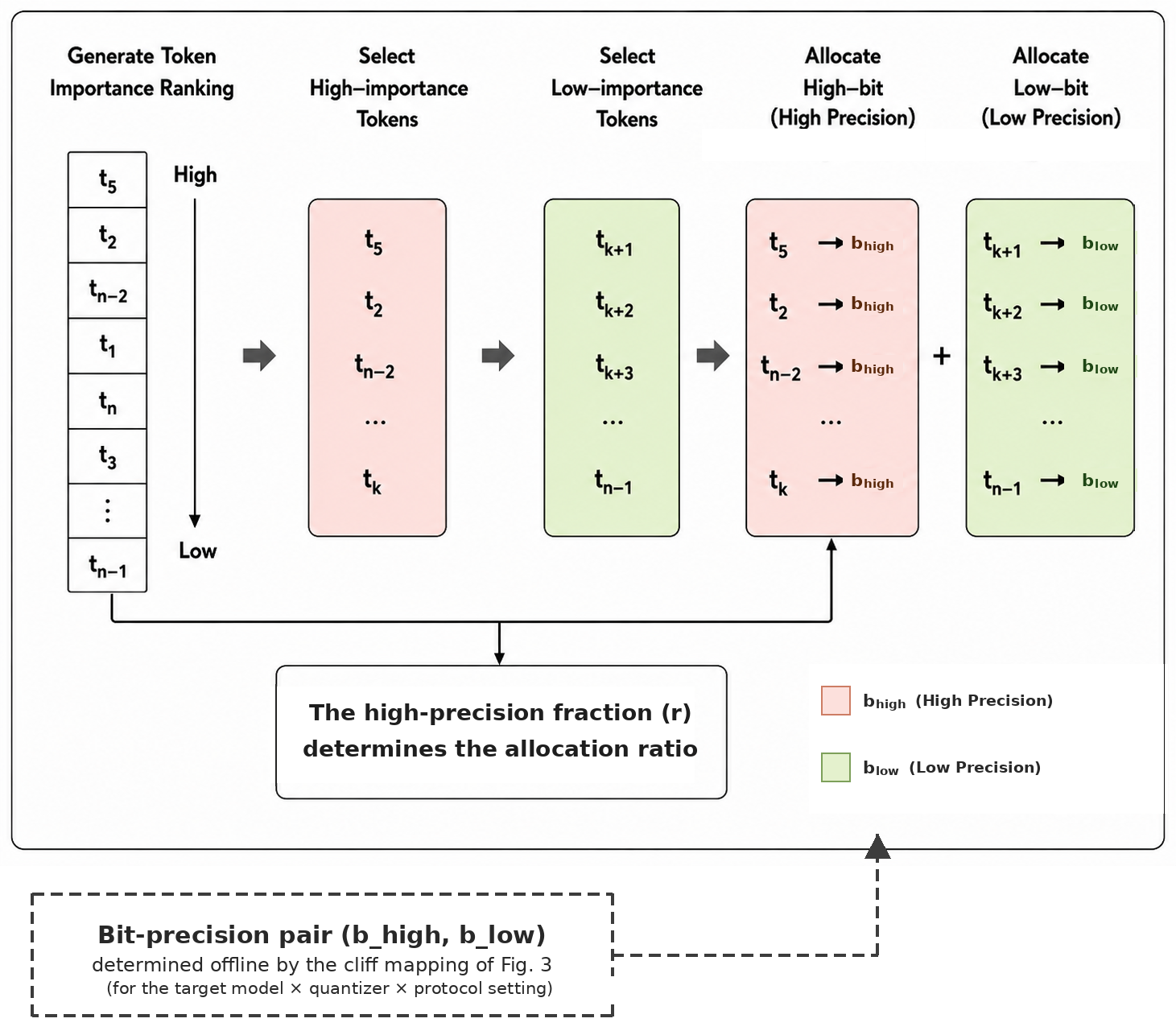}
\caption{Adaptive bit allocation based on token importance. The top-ranked tokens are assigned high precision, while the remaining tokens are assigned low precision. The dashed box indicates that the bit-precision pair $(b_{high}, b_{low})$ is supplied by the offline cliff mapping of Fig.~3 and revalidated when the target model, quantizer, protocol, or context-budget composition changes materially. The allocation is realized purely by the top-$r$ ranking; no score threshold is estimated. Token subscripts in the two groups denote membership in $\mathcal{H}$ and $\mathcal{I}\setminus\mathcal{H}$ respectively; the low-precision group is generally not contiguous in token position.}
\label{fig:allocation}
\end{figure}
\FloatBarrier
Once the semantic importance score $s_i$ is computed, SemKV selects the top $r$ fraction of all tokens as high-precision tokens.
\begin{equation}
K = \lfloor rT \rfloor,\quad
\mathcal{H} = \mathrm{TopK}(\{s_i\}_{i=1}^{T}, K).
\end{equation}

The bit-width $b_i$ assigned to token $i$ is then defined as follows.
\begin{equation}
b_i =
\begin{cases}
b_{\mathrm{high}}, & \text{if } i \in \mathcal{H}, \\
b_{\mathrm{low}}, & \text{otherwise}.
\end{cases}
\end{equation}

The average bit-width is given as follows.
\begin{equation}
\bar{b} = \frac{1}{T}\sum_{i=1}^{T} b_i .
\end{equation}

\subsection{KV Cache Quantization}

\begin{figure}[H]
\centering
\includegraphics[width=\textwidth]{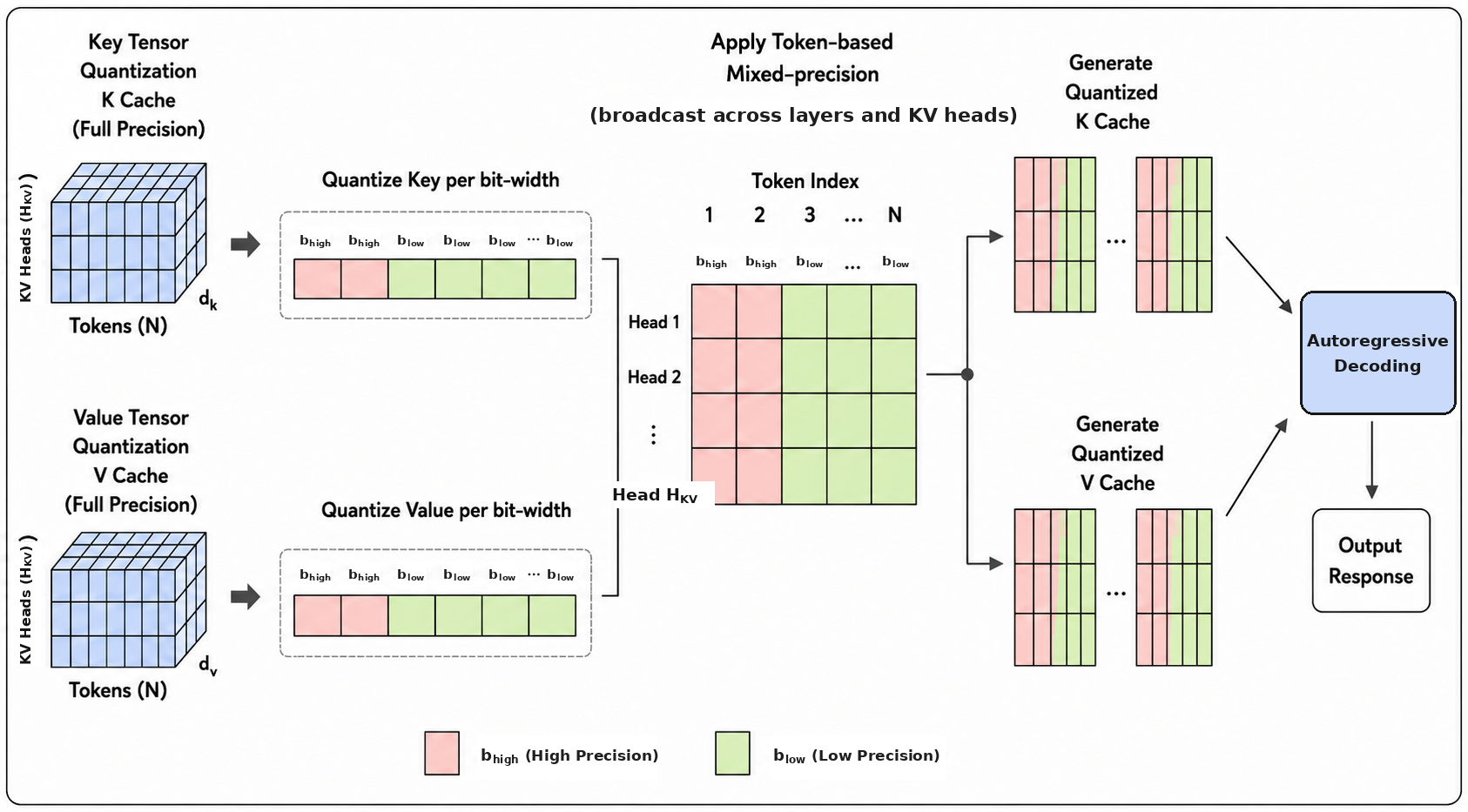}
\caption{Mixed-precision KV quantization process. The token-level bit allocation is applied to key and value tensors, resulting in a mixed-precision quantized KV cache used for decoding. The token-wise precision assignment is broadcast across all layers and KV heads ($H_{\mathrm{KV}}$); bits are never allocated along the head dimension itself. The $b_{\mathrm{high}}$/$b_{\mathrm{low}}$ labels inside the panel denote the two precision levels; experiments use fractional-level codes such as 2.585/2.322 bits (Sec.~4.1).}
\label{fig:kv_quant}
\end{figure}
\FloatBarrier
Let the key and value of token $i$ at layer $l$ be $K_i^{(l)}$ and $V_i^{(l)}$, respectively. Then the quantized KV cache of SemKV can be expressed as follows.
\begin{equation}
\hat{K}_i^{(l)} = Q_{b_i}(K_i^{(l)}),\quad
\hat{V}_i^{(l)} = Q_{b_i}(V_i^{(l)}).
\end{equation}

\subsection{All-Token-Preserving Property}

Pruning-based methods retain only a proper subset $\mathcal{P}$ of the input-token set $\mathcal{I}$. In contrast, SemKV preserves the complete token set and changes only the precision assigned to each token.
\begin{equation}
\mathcal{P} \subset \mathcal{I},\quad |\mathcal{P}| < |\mathcal{I}|.
\end{equation}

SemKV instead satisfies the following condition.
\begin{equation}
\mathcal{I}_{\mathrm{SemKV}} = \mathcal{I}.
\end{equation}

That is, SemKV keeps every token in the input prompt in the KV cache. The only difference lies in the precision assigned to each token.

\subsection{Generation-Token Mixed-Precision Quantization}

The mechanism above quantizes the prompt KV cache once after prefill, when the full score distribution is available and an exact top-$r$ selection is possible. Generation-time tokens do not enjoy this property: their scores arrive one by one, and committing a bit-width immediately upon generation requires approximating an unknown future score distribution, which we found brittle in practice. SemKV instead applies a \emph{block-exact deferred} scheme, illustrated in Fig.~6. We call the procedure \emph{block-exact} because the prescribed high-precision count is met exactly within each completed block; it does not claim equivalence to a global top-$r$ ranking over all generated tokens. Newly generated tokens are appended to the cache at full precision while only their importance scores are collected; whenever a block of $B$ tokens (e.g., $B{=}64$) has accumulated, exactly $\operatorname{round}(rB)$ tokens with the highest scores are selected within the block --- where $\operatorname{round}(\cdot)$ denotes nearest-integer rounding --- and only those positions are quantized to $b_{\mathrm{high}}$, the remainder to $b_{\mathrm{low}}$. At the end of each generation turn, the residual partial block is quantized using the same block-wise top-$r$ rule. This guarantees the prescribed high-precision fraction constructively (per-block rounding error at most half a token), removes all threshold estimation, and keeps the most recent fewer than $B$ tokens temporarily uncompressed during decoding, while also retaining the most recent local context at full precision. In multi-turn dialogue the same procedure is applied per turn on the accumulated conversation.

\begin{figure}[H]
\centering
\includegraphics[width=\textwidth]{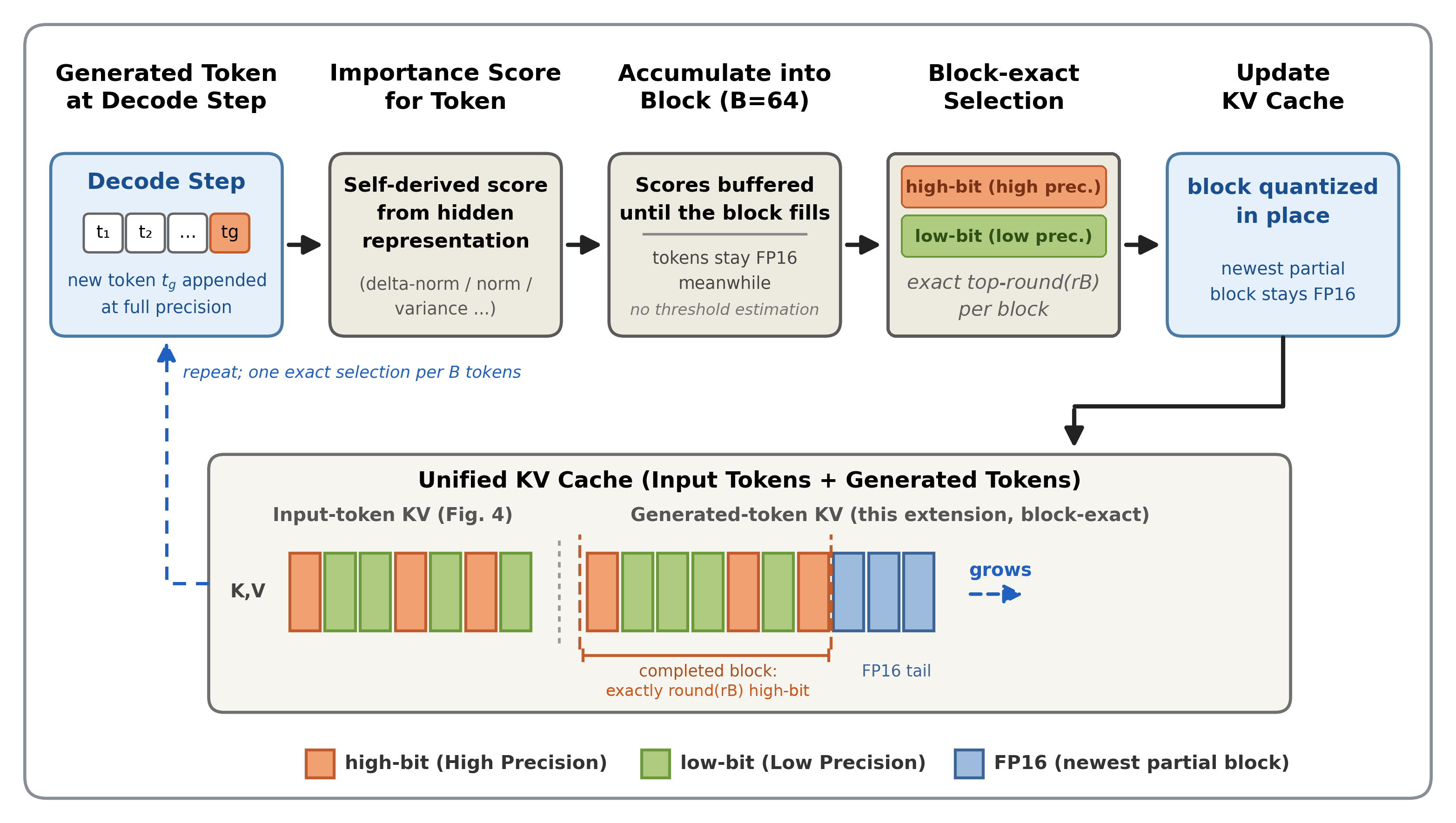}
\caption{Generation-token mixed-precision quantization. Scores of newly generated tokens are accumulated and quantized block-exactly in deferred fashion, so the prescribed high-precision fraction is met constructively without threshold estimation, and the most recent partial block remains in FP16 during decoding; at the end of each generation turn, the residual block is quantized using the same block-wise top-$r$ rule.}
\end{figure}
\FloatBarrier

\subsection{SemKV Algorithm}

\begin{algorithm}[H]
\caption{SemKV Semantic Mixed-Precision KV Cache Compression (prefill)}
\begin{algorithmic}[1]
\REQUIRE Prompt $X=\{x_1,\ldots,x_T\}$, LLM $f_\theta$, high-precision fraction $r$, low bit $b_{\mathrm{low}}$, high bit $b_{\mathrm{high}}$, scoring layer count $m$
\ENSURE Mixed-precision quantized KV cache $\hat{\mathcal{C}}_{\mathrm{KV}}$
\STATE Perform prefill forward pass for input prompt $X$.
\STATE Obtain KV cache $\mathcal{C}_{\mathrm{KV}}^{(l)}=\{K^{(l)},V^{(l)}\}$ for each layer $l$.
\STATE Compute semantic importance score $s_i$ using the last $m$ layers.
\STATE Set $K=\lfloor rT \rfloor$ and select $\mathcal{H}=\mathrm{TopK}(\{s_i\}_{i=1}^{T},K)$.
\FOR{each token $i$}
    \IF{$i \in \mathcal{H}$}
        \STATE $b_i \leftarrow b_{\mathrm{high}}$
    \ELSE
        \STATE $b_i \leftarrow b_{\mathrm{low}}$
    \ENDIF
\ENDFOR
\STATE Quantize $K_i^{(l)}$ and $V_i^{(l)}$ using $Q_{b_i}(\cdot)$.
\STATE Use $\hat{\mathcal{C}}_{\mathrm{KV}}$ for decoding; generated tokens are handled by Algorithm~2.
\end{algorithmic}
\end{algorithm}

\begin{algorithm}[H]
\caption{Block-Exact Quantization of Generated Tokens (Sec.~3.7)}
\begin{algorithmic}[1]
\REQUIRE Quantized prompt cache $\hat{\mathcal{C}}_{\mathrm{KV}}$ from Algorithm~1, high-precision fraction $r$, bits $b_{\mathrm{high}}, b_{\mathrm{low}}$, block size $B$
\STATE Initialize score buffer $\mathcal{S}\leftarrow\emptyset$ (scores stored with their token indices).
\WHILE{generating}
    \STATE Generate token $t_g$; append $K_g^{(l)},V_g^{(l)}$ to $\hat{\mathcal{C}}_{\mathrm{KV}}$ at full precision.
    \STATE Compute score $s_g$ with the same indicator; $\mathcal{S}\leftarrow\mathcal{S}\cup\{(g,s_g)\}$. \COMMENT{no per-token decision}
    \IF{$|\mathcal{S}| = B$}
        \STATE $\mathcal{H}_{\mathrm{blk}}\leftarrow\mathrm{TopKIndices}(\{s_g:(g,s_g)\in\mathcal{S}\},\ \operatorname{round}(rB))$ \COMMENT{exact count within the block}
        \STATE For each token $g$ in the block: $b_g \leftarrow b_{\mathrm{high}}$ if $g\in\mathcal{H}_{\mathrm{blk}}$ else $b_{\mathrm{low}}$.
        \STATE Quantize the block's $K_g^{(l)},V_g^{(l)}$ in place with $Q_{b_g}(\cdot)$; $\mathcal{S}\leftarrow\emptyset$.
    \ENDIF
\ENDWHILE
\STATE During decoding, the most recent partial block ($<B$ tokens) stays at full precision.
\STATE At the end of each generation turn, if $|\mathcal{S}|>0$, flush the residual partial block with the same rule: $\mathcal{H}_{\mathrm{blk}}\leftarrow\mathrm{TopKIndices}(\{s_g:(g,s_g)\in\mathcal{S}\},\ \operatorname{round}(r|\mathcal{S}|))$, then quantize in place.
\end{algorithmic}
\end{algorithm}

\FloatBarrier

\section{Experiments and Results}

This section evaluates SemKV along the narrative established above: (i) the quality cliff of uniform KV quantization on a fractional-bit grid, (ii) grid interpolation by all-token-preserving mixed precision and its indicator ablation, (iii) generation-token quantization, (iv) scoring cost, (v) multi-turn behavior, (vi) deletion versus low-precision preservation, (vii) a below-cliff stress test, and (viii) cross-model and context-length transfer.

\subsection{Experimental Setup and Statistical Protocol}

\textbf{Models and benchmarks.} The main backbone is Llama-3.1-8B-Instruct; transfer experiments use Mistral-7B-Instruct-v0.3. Single-turn evaluation follows our LongBench protocol on ten subsets (qasper, multifieldqa\_en, hotpotqa, 2wikimqa, gov\_report, multi\_news, triviaqa, samsum, passage\_retrieval\_en, lcc) with 30 samples per subset and a 7{,}500-token context budget. The composite score is the item-average of each task's official metric (QA-F1, ROUGE, accuracy, or code similarity, as specified by LongBench), not QA-F1 alone. Multi-turn evaluation uses MT-Eval~[80] with ROUGE-L per turn.

\textbf{Quantization.} All quantized settings use integer-level scalar quantization with a Hadamard rotation; fractional bit-widths denote non-power-of-two level counts (e.g., $2.322 = \log_2 5$, $2.585 = \log_2 6$ bits). SemKV uses two-level allocation between adjacent above-cliff grid points, $b_{\mathrm{high}}{=}2.585$ and $b_{\mathrm{low}}{=}2.322$, with a high-precision fraction of $r{=}25\%$ for single-turn prompts (average 2.39 bits) and $r{=}35\%$ including generation tokens in multi-turn settings (average 2.41 bits).

\textbf{Memory accounting.} Reported bit-widths are code bits; storage adds two components. First, non-power-of-two level counts are stored by mixed-radix packing: three five-level codes pack into 7 bits ($2.333$ bits/value, $+0.49\%$ over $\log_2 5$) and five six-level codes into 13 bits ($2.600$ bits/value, $+0.58\%$ over $\log_2 6$). Second, the per-row asymmetric affine quantizer stores two FP16 parameters (offset and scale) per head--token row of dimension 128, adding $0.25$ bits/value; the per-token precision flag contributes 1 bit per token ($<10^{-4}$ bits/value) and is negligible. The effective storage cost of the 2.39-bit operating point is therefore $2.65$ bits/value, i.e., a $6.04\times$ reduction versus FP16. Quality experiments use fake quantization (quantize--dequantize with FP16 compute), so these storage figures are validated separately by a packed-buffer microbenchmark that allocates the exact packed sizes on GPU (Sec.~4.5); measured allocations match this accounting to within allocator page rounding ($<0.9\%$), and the compression ratio is independent of context length. The reported packed-storage ratios exclude the transient FP16 tail of fewer than $B$ generated tokens (Sec.~3.7); this overhead is bounded by $B{-}1$ tokens and becomes negligible for long contexts.

\textbf{Statistical conventions.} Throughout the experiments, $\Delta$ denotes the paired score difference relative to the full-precision (FP16) baseline; $\Delta{=}0$ indicates equal measured scores. Parenthetical values are $p$-values from the stated paired test, and we use $p{<}0.05$ as evidence of a statistically detectable difference. \emph{n.s.}\ means that the test did not detect a significant difference; it does not establish equality. Where shown, error bars represent 95\% confidence intervals of the paired difference. Bonferroni correction is applied to prespecified families of multiple comparisons and is noted where used.

\textbf{Statistical protocol (prespecified).} Every headline condition is run with three seeds (42/43/44); exploratory sweeps use seed 42. The primary test is a paired $t$-test over items keyed by (seed, item, task); the eight indicator-vs-full-KV tests and the 28 pairwise indicator tests are corrected as separate Bonferroni families. Single-seed exploratory findings were not treated as confirmed effects unless their direction was consistent across all three seeds. Table~2 summarizes the configuration.

Three conventions matter for interpreting what follows. First, at $n{=}900$ the paired 95\% CI of the LongBench composite is approximately $\pm0.011$, so absolute deficits substantially smaller than about $0.011$ cannot be resolved under this protocol and are reported as n.s.\ rather than as absence of loss; \emph{safe} accordingly means that no statistically detectable deficit is observed under the stated protocol, at this resolution (Sec.~4.9 gives a concrete instance in which a more sensitive protocol resolves a deficit that is n.s.\ here). Second, the \emph{measured cliff bracket} is the interval between two adjacent tested grid points across which degradation increases sharply, with the lower grid point showing a large statistically detectable deficit and the direction consistent across seeds; a grid point above the cliff may still exhibit a smaller residual deficit --- in particular in full-cache multi-turn evaluation (Sec.~4.6) --- so lying above the collapse cliff does not by itself imply the absence of any deficit, and the cliff claim rests on the contrast between a flat region of n.s.\ differences and an adjacent collapse significant by several orders of magnitude. Third, because each benchmark item is evaluated under three seeds, the $n{=}900$ paired samples share item-level structure; we therefore verified the headline single-turn and indicator-ablation conclusions using item-level seed averages ($n{=}300$ paired items), and the significance pattern is unchanged (SemKV at 2.39 bits n.s., $p{=}0.85$; uniform 2.585/2.322 bits n.s., $p{=}0.07$/$0.21$; the 2.0-bit collapse significant, $p{<}10^{-6}$; all eight indicators n.s.\ vs.\ full KV and all 28 pairwise comparisons n.s.\ after correction).

\begin{table}[H]
\centering
\caption{SemKV experimental configuration.}
\small
\begin{tabular}{ll}
\toprule
Configuration & Value \\
\midrule
Backbone models & Llama-3.1-8B-Instruct / Mistral-7B-Instruct-v0.3 \\
Scoring (default) & Hidden-state delta norm, last 4 layers \\
Indicator ablation & 8 model-internal indicators (4 families) \\
Precision pair & $b_{\mathrm{high}}{=}2.585$, $b_{\mathrm{low}}{=}2.322$ (above-cliff grid points) \\
High-precision fraction & 25\% (single-turn) / 35\% (multi-turn incl. generation) \\
Quantizer & Integer-level scalar quantization + Hadamard rotation \\
Base-quantizer transfer & TurboQuant-MSE [9] (Sec.~4.9) \\
Generation scheme & Block-exact deferred top-$r$ ($B{=}64$) \\
Storage accounting & Mixed-radix packing + 0.25 bits/value metadata \\
Evaluation & Our LongBench protocol (official task-specific metrics) \\
 & / MT-Eval ROUGE-L \\
Statistics & 3 seeds; paired $t$-test (primary); Bonferroni correction; \\
 & \quad item-level seed-average robustness check \\
\bottomrule
\end{tabular}
\end{table}

\subsection{The Quality Cliff of Uniform KV Quantization}
\label{sec:cliff}

Sweeping uniform quantization over the fractional-bit grid reveals that quality does not decay smoothly (Fig.~7a). On Llama-3.1-8B, every uniform setting in $[2.322, 3.0]$ bits --- including the fractional points $2.807=\log_2 7$, $2.585=\log_2 6$, and $2.322=\log_2 5$ --- is statistically indistinguishable from full KV under the paired test ($|\Delta|\le 0.0094$, all $p\ge 0.09$, $n{=}900$), whereas uniform 2.0 bits collapses ($\Delta{=}-0.0448$, $p{<}10^{-7}$), with the direction consistent in all three seeds. The cliff is therefore located in $(2.0, 2.322]$, and under this base the \emph{same collapse interval} reappears in generation-token quantization (Sec.~4.4) and in multi-turn dialogue (Sec.~4.6, where a small residual above-cliff floor additionally exists), and transfers to Mistral (Sec.~4.9). Cliff \emph{depth}, unlike location, is seed-dependent; all cliff claims are pooled three-seed results.

\textbf{The affine cliff interior is not safely navigable under the tested mixtures.} Integer-level scalar quantization admits \emph{no uniform grid point} strictly between 2.0 bits (4 levels) and 2.322 bits (5 levels); the only way to probe the interior is mixing. Fig.~7b reports three-seed interior probes that mix the two boundary precisions (2.322/2.0) at protection ratios $r\in\{25, 50, 75\}\%$, i.e., average 2.081/2.161/2.242 bits. Recovery is monotone in $r$ but strikingly slow: protecting 25\% at 2.322 bits leaves quality at the u2.0 collapse level ($\Delta{=}-0.0446$ vs.\ $-0.0448$), and even 75\% protection remains significantly degraded ($\Delta{=}-0.0266$, $p{<}10^{-4}$). Moreover, inside the cliff, importance-concentrated protection \emph{underperforms} spread-out protection: delta-norm selection at $r{=}50\%$ is significantly worse than random selection at the same budget ($\Delta{=}-0.0189$, $p{<}0.01$, three-seed consistent). Below the cliff, degradation is broad-based rather than confined to a small evidence set, so concentrating the protection budget on high-scoring tokens systematically exposes the rest. Both observations sharpen the operating rule that motivates SemKV: mixed precision should \emph{interpolate strictly between above-cliff grid points}, not average across the cliff.

\begin{figure}[H]
\centering
\includegraphics[width=\textwidth]{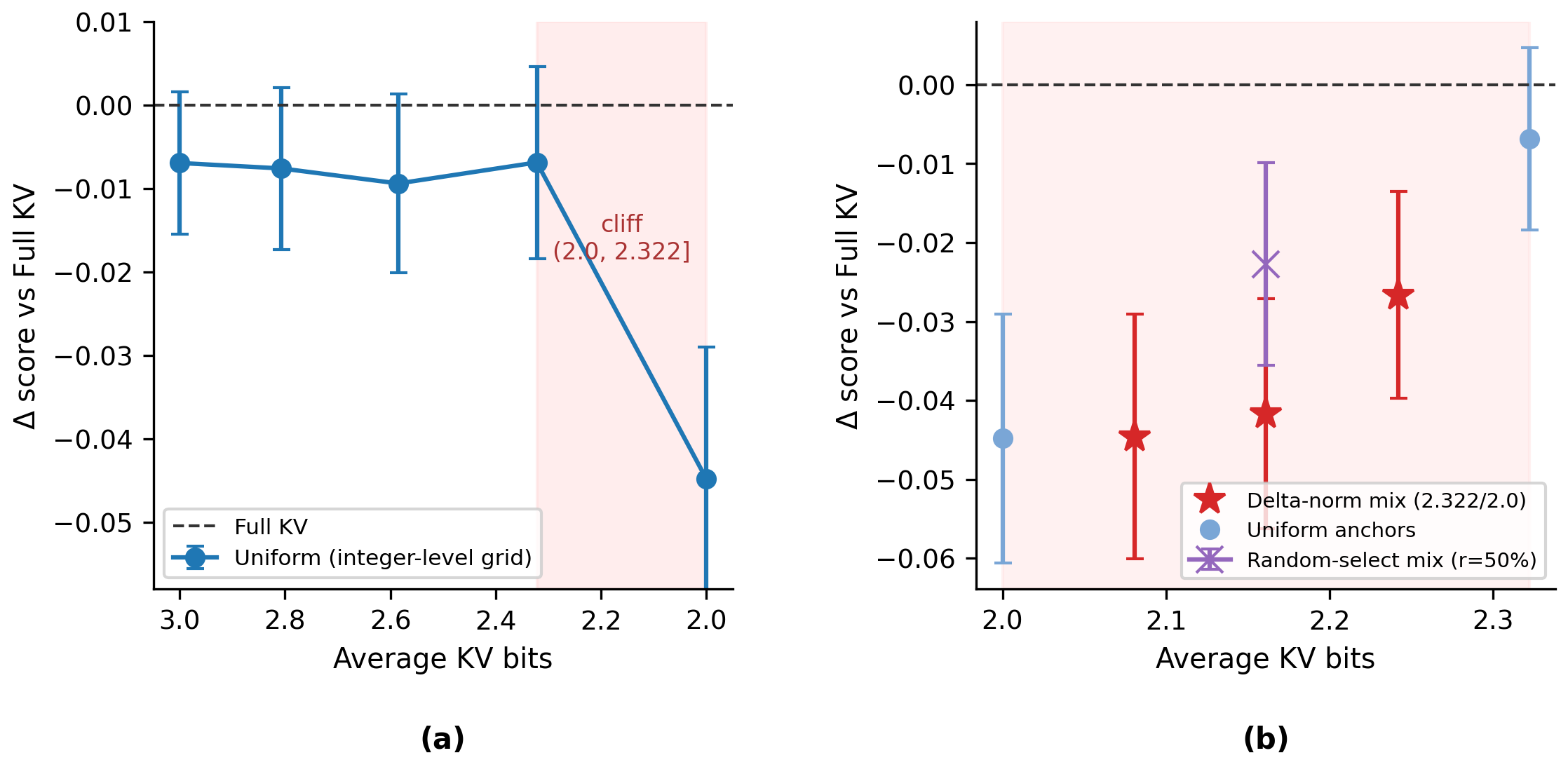}
\caption{The quality cliff (Llama-3.1-8B, LongBench protocol, 3 seeds; error bars: 95\% CI of the paired difference). (a) All uniform settings in $[2.322, 3.0]$ bits are statistically indistinguishable from full KV under this protocol; 2.0 bits collapses. (b) Interior probes mixing 2.322/2.0 bits: recovery is monotone but even 75\% protection remains significantly degraded, and importance-concentrated selection underperforms spread-out selection inside the cliff.}
\end{figure}

\subsection{Grid Interpolation Above the Cliff and Indicator Indifference}
\label{sec:grid}

The cliff delimits where mixed precision is useful: uniform quantization can only occupy grid points, while ranking-based mixed precision interpolates between them. Fig.~8 and Table~3 show the interpolation result at 2.39 average bits (25\% at 2.585, 75\% at 2.322) on our LongBench protocol. Across \emph{eight} model-internal indicators spanning four signal families --- hidden-state (delta norm, norm, variance, cosine delta), KV-norm (key norm, value norm), logit (self-information), and attention (accumulated attention) --- every SemKV variant is statistically indistinguishable from full KV, and all pairwise indicator comparisons are non-significant after Bonferroni correction. In the flat regime the benefit is thus \emph{structural}: preserving all tokens at above-cliff precisions suffices, and at this Llama operating point indicator choice among the tested model-internal signals does not measurably affect quality. This motivates choosing the indicator by \emph{cost} (Sec.~4.5) rather than by quality.

\begin{figure}[H]
\centering
\includegraphics[width=0.9\linewidth]{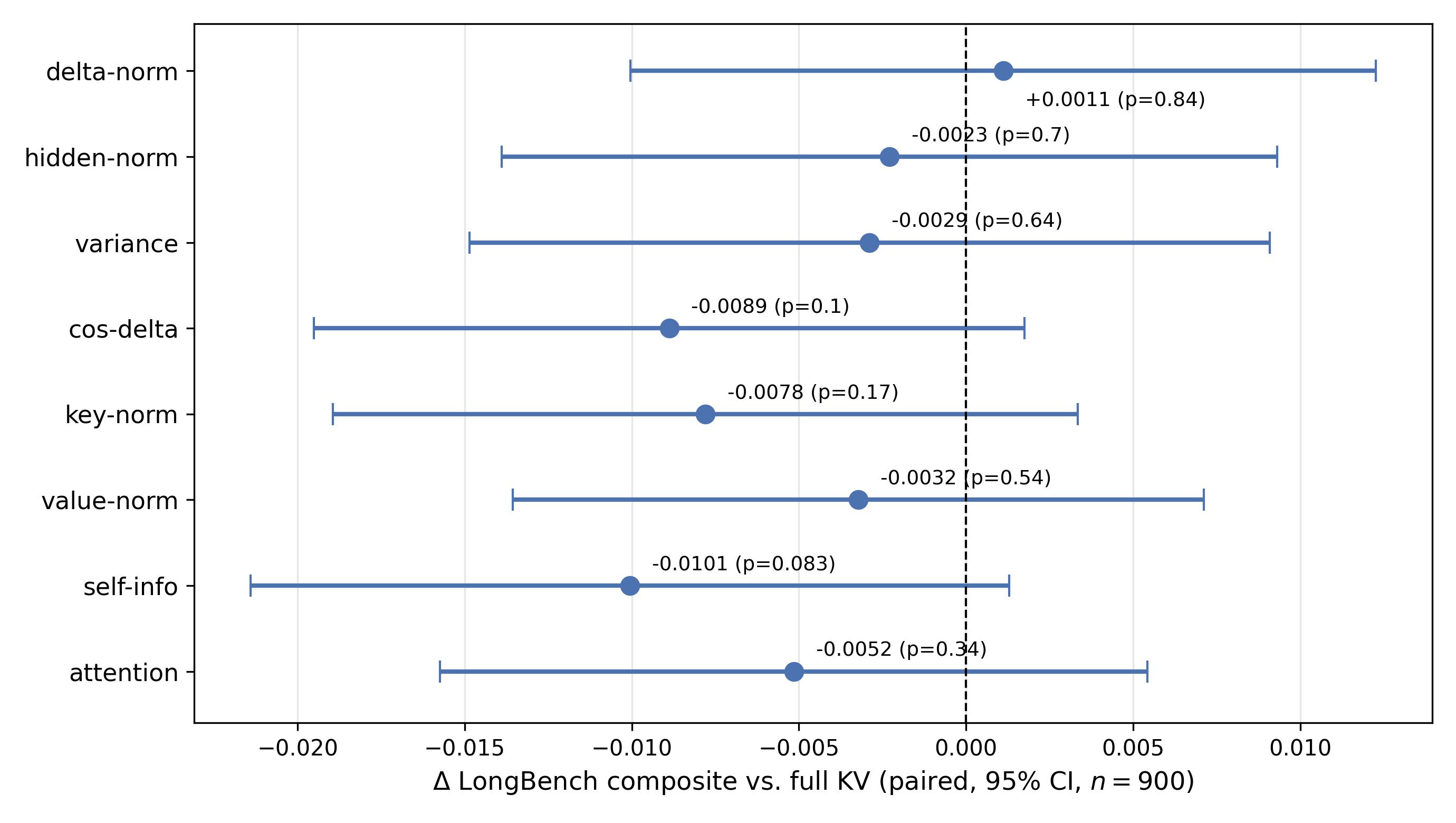}
\caption{Grid interpolation at 2.39 average bits on our LongBench protocol (Llama-3.1-8B, 10 subsets $\times$ 30 items $\times$ 3 seeds). Paired difference from full KV with 95\% confidence intervals for all eight indicator variants of SemKV at the 2.39-bit operating point ($n{=}900$, 3 seeds). Every interval contains zero, and all 28 pairwise indicator comparisons are n.s.\ (Bonferroni-corrected) --- interpolation quality at this operating point is indicator-independent.}
\end{figure}

\begin{table}[H]
\centering
\caption{Grid interpolation at 2.39 bits: paired difference vs.\ full KV (Llama-3.1-8B, $n{=}900$). No variant differs significantly from full KV; all pairwise indicator differences are n.s.\ (Bonferroni-corrected). Compression is measured KV storage vs.\ FP16, metadata included (Sec.~4.1).}
\small
\resizebox{\textwidth}{!}{%
\begin{tabular}{lcccc}
\toprule
Method & Nominal avg.\ code bits & Compression & Score & $\Delta$ vs.\ full ($p$) \\
\midrule
Full KV & 16.0 & $1.0\times$ & 0.4406 & --- \\
Uniform 2.585-bit & 2.585 & $5.6\times$ & 0.4312 & $-0.0094$ (0.09) \\
SemKV hidden-norm & 2.39 & $6.0\times$ & 0.4383 & $-0.0023$ (0.70) \\
SemKV variance & 2.39 & $6.0\times$ & 0.4377 & $-0.0029$ (0.64) \\
SemKV cos-delta & 2.39 & $6.0\times$ & 0.4317 & $-0.0089$ (0.10) \\
SemKV key-norm & 2.39 & $6.0\times$ & 0.4328 & $-0.0078$ (0.17) \\
SemKV value-norm & 2.39 & $6.0\times$ & 0.4374 & $-0.0032$ (0.54) \\
SemKV self-info & 2.39 & $6.0\times$ & 0.4305 & $-0.0101$ (0.08) \\
SemKV attention & 2.39 & $6.0\times$ & 0.4355 & $-0.0052$ (0.34) \\
SemKV delta-norm & 2.39 & $6.0\times$ & 0.4417 & $+0.0011$ (0.84) \\
\bottomrule
\end{tabular}}%

\end{table}

\subsection{Generation-Token Quantization: Affine-Base Collapse and a Favorable Operating Point}
\label{sec:gen}

Applying the block-exact scheme of Sec.~3.7 to generation tokens reproduces the cliff on the generation side (Fig.~9). Quantizing generated tokens at 2.322 bits is a favorable operating point: quality is statistically indistinguishable from FP16 generation ($\Delta{=}-0.0041$, $p{=}0.14$, $n{=}900$), and the max-length termination rate is significantly \emph{reduced} relative to FP16 (McNemar $p{<}10^{-4}$). In contrast, any configuration whose low-bit side is 2.0 --- uniform 2.0-bit generation or mixed generation with $b_{\mathrm{low}}{=}2.0$, even when paired with 2.585- or 3.0-bit high precision --- degrades significantly ($\Delta$ between $-0.0127$ and $-0.0157$, all $p{<}10^{-3}$, three-seed consistent) and elevates the max-length termination rate (McNemar $p{=}0.015$ for uniform 2.0). The asymmetry is the generation-side signature of the cliff: a below-cliff component cannot be compensated by mixing in higher precision, mirroring the interior result of Sec.~4.2.

\begin{figure}[H]
\centering
\includegraphics[width=\textwidth]{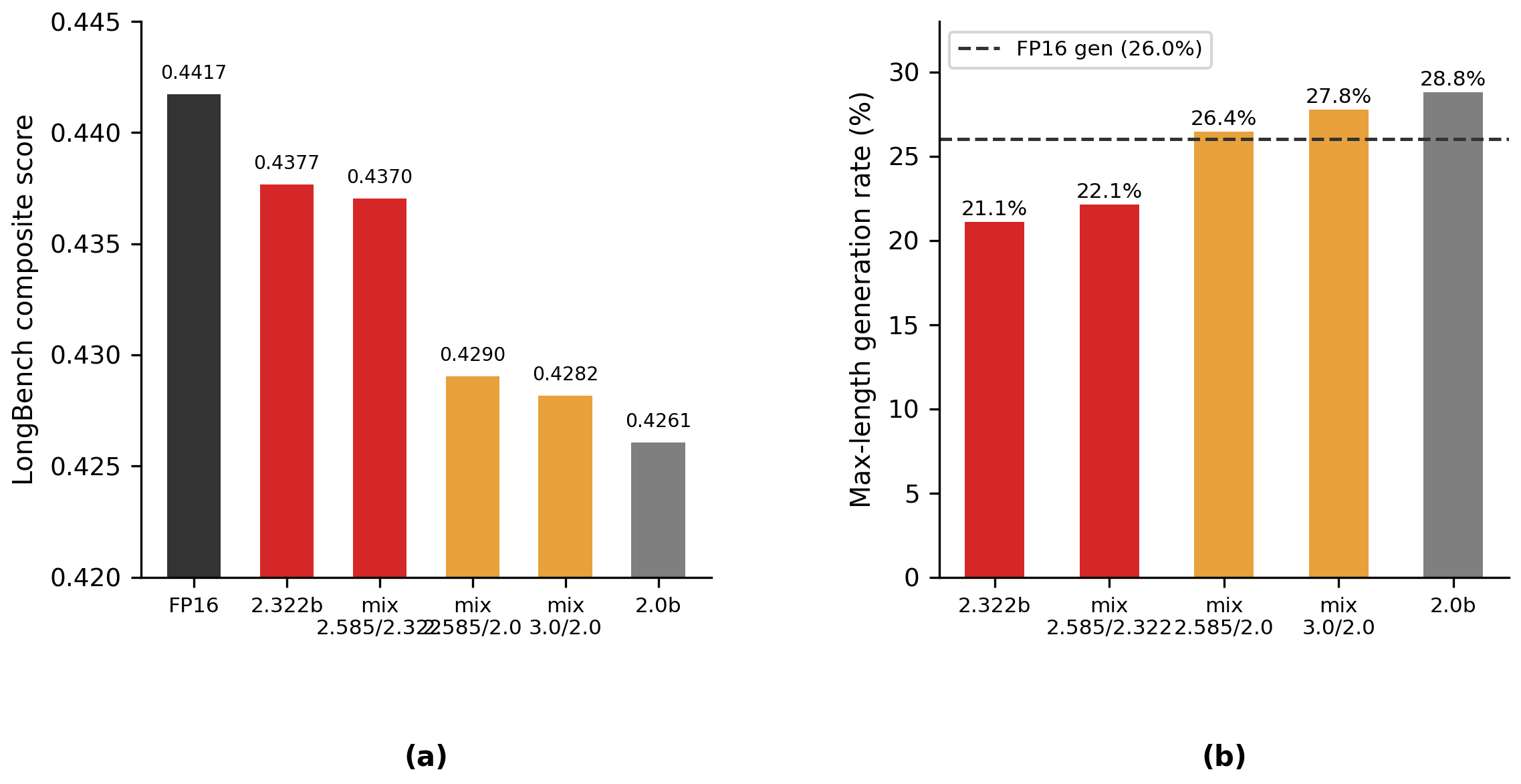}
\caption{Generation-token quantization (Llama-3.1-8B, 3 seeds). (a) 2.322-bit generation is statistically indistinguishable from FP16 generation; every configuration containing a 2.0-bit generation component degrades significantly, regardless of the paired high precision. (b) The max-length termination rate is reduced at 2.322 bits and elevated only at 2.0 bits. Max-length termination rate: the fraction of outputs that reach the configured generation limit without emitting EOS. We use this rate as an operational proxy for runaway-generation behavior.}
\end{figure}

\subsection{Cost: Scoring Overhead and Memory Footprint}
\label{sec:cost}

At the primary Llama single-turn operating point the eight tested indicators are statistically indistinguishable, so computational cost becomes the practical selection criterion (Fig.~10). Hidden-state scoring is $O(1)$ per decoding step and essentially flat in context length ($+3.1\%$ from 1k to 32k tokens in our microbenchmark), whereas attention-score recomputation is $O(T)$ per step, growing to $3.2\times$ the 1k cost by 32k and $8.3\times$ the hidden-state cost at that length. Attention scoring additionally requires materializing attention weights (eager attention), itself a $\sim$5\% decoding-throughput penalty in our measurement. On the LongBench workload the end-to-end scoring share of generation time was 0.39\% for hidden-state scoring versus 1.11\% for attention recomputation ($2.9\times$). Across the tested model-internal indicators, a per-prompt profiling of the prefill scoring pass shows a further spread: value-norm, which reads the cached value tensors directly, costs 0.2\,ms and 22\,MiB of peak activation memory per prompt, versus 1.1--1.2\,ms / 263\,MiB for hidden-norm and variance, 3.0\,ms / 350\,MiB for delta-norm, and 3.3\,ms / 438\,MiB for cosine-delta --- value-norm is roughly $15\times$ cheaper in time and $16\times$ lower in peak activation memory than delta-norm because it avoids materializing multi-layer hidden states. Since the eight indicators are quality-indistinguishable at this operating point, value-norm is the lowest-cost indicator among those profiled in this comparison (the four hidden-state indicators and value-norm) when scoring memory is at a premium. SemKV nonetheless keeps delta-norm as the reference configuration throughout this paper: it is a conservative, relatively high-cost model-internal indicator (although cosine-delta incurs slightly higher scoring time and peak activation memory), so the reported reference overhead represents a conservative high-cost configuration, though not the absolute maximum among all tested indicators.

\begin{figure}[H]
\centering
\includegraphics[width=\textwidth]{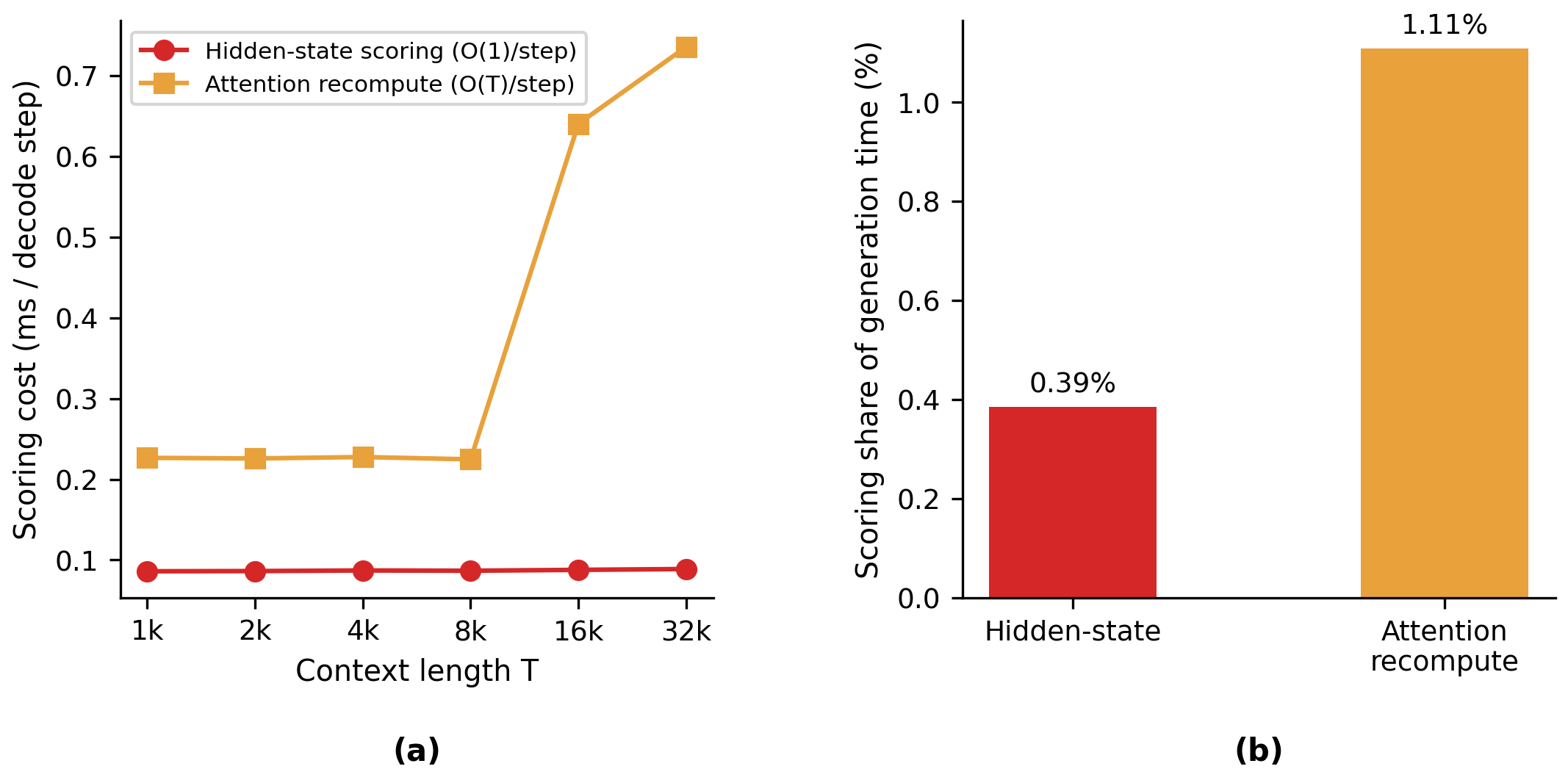}
\caption{Scoring cost (Llama-3.1-8B). (a) Per-step scoring cost versus context length: hidden-state scoring is flat ($O(1)$), attention recomputation grows with $T$. (b) End-to-end scoring share of generation time on LongBench. Complexity labels are with respect to context length $T$: hidden-state scoring is $O(1)$ per decode step in $T$, while attention recomputation grows with $T$.}
\end{figure}

\textbf{Measured memory footprint.} Fig.~11 reports the packed-storage microbenchmark of Sec.~4.1: KV-shaped buffers at the exact packed sizes (code bits plus metadata) are allocated on GPU and measured directly, with the FP16 baseline allocated as per-layer K/V tensors. Both backbones share the KV shape (32 layers, 8 KV heads, head dimension 128), so one measurement covers both. Across context lengths from 4k to 128k tokens, SemKV at the 2.39-bit operating point occupies $6.0\times$ less storage than full FP16 KV in every measurement ($6.02$--$6.04\times$; e.g., 16.0\,GiB $\rightarrow$ 2.65\,GiB at 128k), the multi-turn 2.41-bit point is $6.0\times$ as well, and the measured ratio is constant in context length, as the accounting predicts. The FP16 keep-25\% pruning baseline of Sec.~4.7 measures $4.0\times$, so SemKV outperforms it while using a $1.5\times$ smaller footprint. Measured allocations agree with the analytical accounting to within CUDA allocator page rounding ($<0.9\%$, largest at the shortest context).

\begin{figure}[H]
\centering
\includegraphics[width=\textwidth]{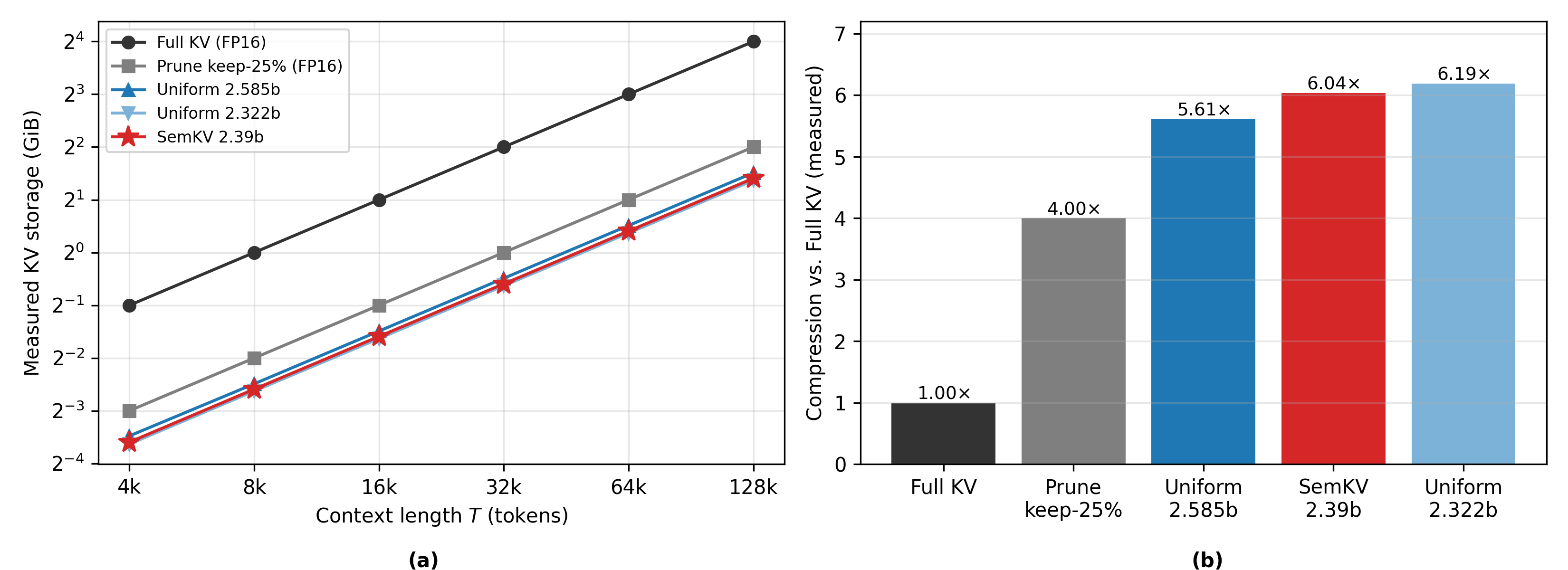}
\caption{Measured KV storage (packed-buffer microbenchmark; both backbones share this KV shape). (a) KV storage versus context length: SemKV at 2.39 bits tracks $6.0\times$ below full FP16 KV at every length. (b) Measured compression at 128k tokens; the pruning baseline that SemKV outperforms in Sec.~4.7 occupies a $1.5\times$ larger footprint.}
\end{figure}

\subsection{Multi-Turn Dialogue}
\label{sec:multiturn}

On MT-Eval, catastrophic degradation again appears at a single grid step: only uniform 2.0-bit collapses ($\Delta{=}-0.082$, $p{<}10^{-6}$), and collapse depth tracks how strongly a task depends on prior content (expansion $-0.181$ $>$ follow-up $-0.109$ $>$ refinement $-0.067$ $>$ recollection $-0.024$, n.s.). Above the cliff the multi-turn picture is not perfectly flat: with adequate statistical power, full-cache multi-turn quantization leaves a small but significant residual deficit under this affine base --- uniform 2.322-bit at $-0.021$ in both the three-seed follow-up run ($p{<}10^{-3}$, $n{=}90$) and the original five-task affine run ($p{<}10^{-3}$, $n{=}80$), with uniform 2.585/2.807 similar ($-0.012$/$-0.015$), i.e.\ the deficit is insensitive to bits well above the cliff. SemKV at 2.41 average bits, with generation tokens quantized block-exactly, sits at $-0.008$ to $-0.009$ versus full KV (borderline, $p{=}0.03$--$0.06$ depending on the run) --- it significantly improves on uniform 2.322-bit ($+0.0137$, $p{=}0.006$) and is statistically indistinguishable from uniform 2.585-bit quality while using $0.17$ fewer code bits, but it does not beat this small protocol floor; Sec.~4.9 shows the same structure, with a $2.5\times$ smaller floor, under the TurboQuant base. The deficit concentrates in the task most dependent on prior generated content (expansion), mirroring the collapse ordering above, and does not track the raw number of generated tokens --- consistent with errors compounding through the dialogue text itself (each turn conditions on previously generated, slightly perturbed answers) rather than with per-step cache reuse. Fig.~12 reports the multi-turn indicator ablation at the operating point. The hidden-state indicators remain close to full KV and are mutually indistinguishable; after Bonferroni correction across indicators none of them differs significantly from full KV, although delta-norm is nominally significant before correction ($\Delta{=}-0.0075$, raw $p{=}0.029$). Attention scoring shows the largest and most consistently detectable deficit ($\Delta{=}-0.0115$, $p{<}0.001$, significant after correction; direct attention-vs-hidden comparisons are n.s.). Together with its higher computational cost (Sec.~4.5), attention scoring offers no observed advantage for generation-time allocation in this setting.

\begin{figure}[H]
\centering
\includegraphics[width=0.75\linewidth]{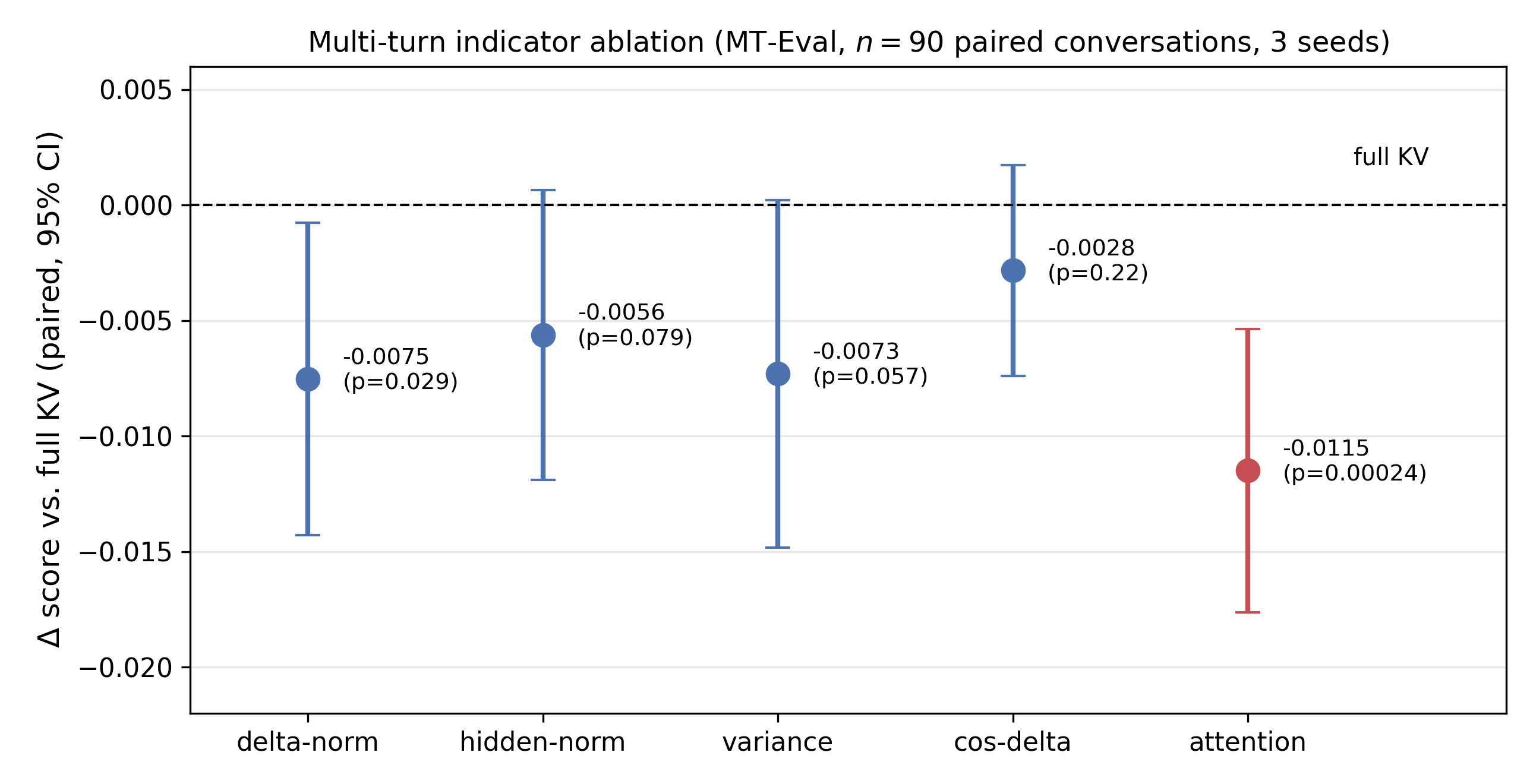}
\caption{Multi-turn indicator ablation (MT-Eval follow-up, $n{=}90$ paired conversations, 3 seeds): paired difference vs.\ full KV with 95\% CIs. Hidden-state indicators are mutually indistinguishable, and after correction none differs significantly from full KV. These hidden-state indicators lie near the small multi-turn floor of this protocol (Secs.~4.6, 4.9). Attention scoring exhibits a larger deficit that remains statistically detectable after correction.}
\end{figure}

\subsection{Deletion versus Low-Precision Preservation}
\label{sec:prune}

Fig.~13 and Table~4 use hidden-norm for the primary \emph{matched-indicator} comparison between pruning and SemKV --- isolating the structural variable (delete vs.\ preserve-at-low-precision) --- and include additional pruning indicators as robustness controls for the selection rule. This is a controlled structural comparison, not a reproduction of engineered eviction systems with recency windows (e.g., H2O, SnapKV). FP16 pruning that keeps 25\% of tokens is a 4-bit-equivalent memory budget --- \emph{more} memory than SemKV at 2.39 bits --- yet collapses to 0.063--0.149 depending on the indicator, while SemKV scores 0.432--0.442. Two observations sharpen the conclusion. First, \emph{random} pruning is statistically indistinguishable from indicator-based pruning (paired difference n.s.) and shows the same qualitative collapse, suggesting that deletion itself, rather than selection quality, is the dominant source of failure in this controlled comparison. Second, collapse depth tracks evidence sparsity across tasks (synthetic retrieval and few-shot collapse hardest; summarization and code least), consistent with the evidence-token-loss mechanism. In the earlier full sweep, collapse was also independent of the keep ratio from 15\% to 50\%.

\begin{figure}[H]
\centering
\includegraphics[width=0.95\linewidth]{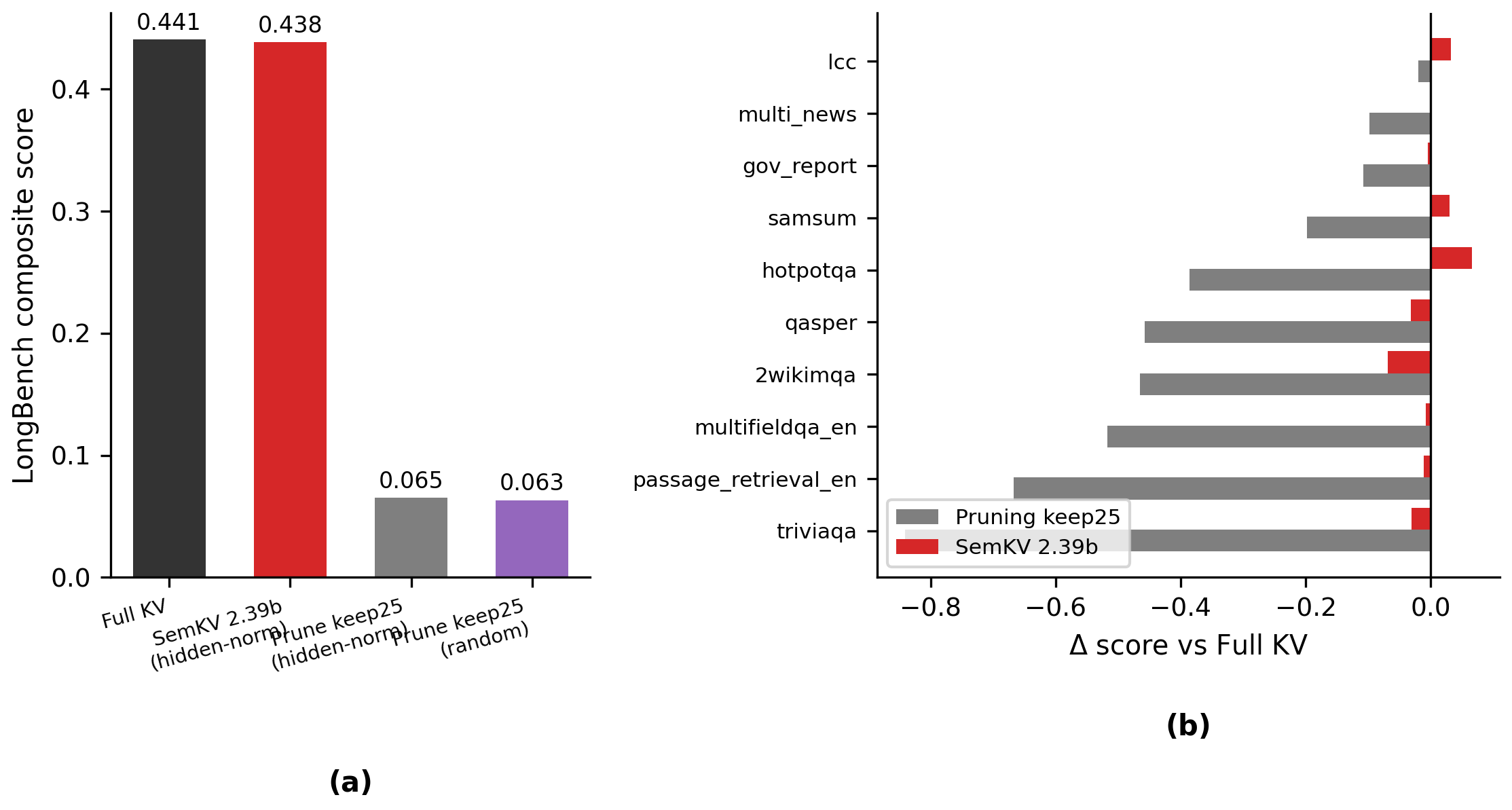}
\caption{Deletion versus low-precision preservation under the same indicator (Llama-3.1-8B, 3 seeds). (a) SemKV at 2.39 bits shows no statistically detectable deficit from full KV while FP16 keep-25\% pruning (a strictly larger, 4-bit-equivalent memory budget) collapses; random pruning is statistically indistinguishable from indicator-based pruning. (b) Per-task collapse depth tracks evidence sparsity. The pruning condition is a controlled structural baseline using the same indicator and budget accounting, not an implementation of an engineered eviction system such as H2O or SnapKV.}
\end{figure}

\begin{table}[H]
\centering
\caption{Structural comparison between SemKV and FP16 pruning (LongBench composite, $n{=}900$). The primary matched-indicator comparison uses hidden-norm; additional pruning indicators test robustness to the selection rule. Compression is measured KV storage vs.\ FP16; SemKV occupies a $1.5\times$ smaller footprint than the pruning baselines it outperforms. For SemKV rows the bit figure is the nominal average code rate on the level grid; for FP16 keep-$k$ pruning it is the memory-equivalent rate $16k$ --- both are converted to measured storage for the compression column.}
\small
\resizebox{\textwidth}{!}{%
\begin{tabular}{lcccc}
\toprule
Method & Nominal / mem.-equiv.\ bits & Compr. & Score & $\Delta$ vs.\ full \\
\midrule
Full KV & 16.0 & $1.0\times$ & 0.4406 & --- \\
SemKV hidden-norm & 2.39 & $6.0\times$ & 0.4383 & $-0.0023$ (0.70) \\
Prune keep-25\% (hidden-norm) & 4.0 & $4.0\times$ & 0.0650 & $-0.3756$ ($p{<}10^{-6}$) \\
Prune keep-25\% (variance) & 4.0 & $4.0\times$ & 0.0676 & $-0.3730$ ($p{<}10^{-6}$) \\
Prune keep-25\% (cos-delta) & 4.0 & $4.0\times$ & 0.1493 & $-0.2913$ ($p{<}10^{-6}$) \\
Prune keep-25\% (random) & 4.0 & $4.0\times$ & 0.0632 & $-0.3774$ ($p{<}10^{-6}$) \\
\bottomrule
\end{tabular}}%

\end{table}

\subsection{Below-Cliff Stress: When Indicator Quality Matters}
\label{sec:stress}

The primary SemKV operating points and indicator ablations above use precision pairs on the high-precision side of the cliff, where the tested indicators are statistically indistinguishable. To probe when selection matters, we stress the system on the task with the deepest multi-turn collapse (MT-Eval expansion) and \emph{widen the precision gap} to 3.0/2.0 bits at $r{=}35\%$ (average 2.30 bits): now 65\% of tokens sit \emph{below} the cliff, so performance therefore depends strongly on which 35\% of tokens receive high precision. Fig.~14 shows the outcome. At the standard operating point (2.585/2.322, average 2.40 bits), SemKV scores 0.394, significantly above uniform 2.322-bit (0.332, $p{<}0.05$) and uniform 2.0-bit (0.254, $p{<}0.01$), while suppressing runaway generation (2/10 conversations vs.\ 4/10 and 7/10). In the widened-gap setting, the model-internal indicators separate into a clear performance spectrum (Fig.~15): value-norm, attention, cosine-delta, self-information, and delta-norm allocate at or above the random-selection baseline, whereas hidden-norm and key-norm fall below it --- i.e., in this regime a bad indicator is worse than no indicator. Given the small stress sample (10 conversations), we report pooled three-seed statistics with per-seed direction consistency and treat individual indicator-vs-random gaps as directional; the value-norm-vs-hidden-norm spread itself is significant ($p{<}0.01$). Although the stress sample is small, the pooled and per-seed trends consistently support the qualitative conclusion: \emph{in the tested Llama setting, structure dominates at the primary above-cliff point, whereas selection becomes decisive under below-cliff stress}.

\begin{figure}[H]
\centering
\includegraphics[width=0.95\linewidth]{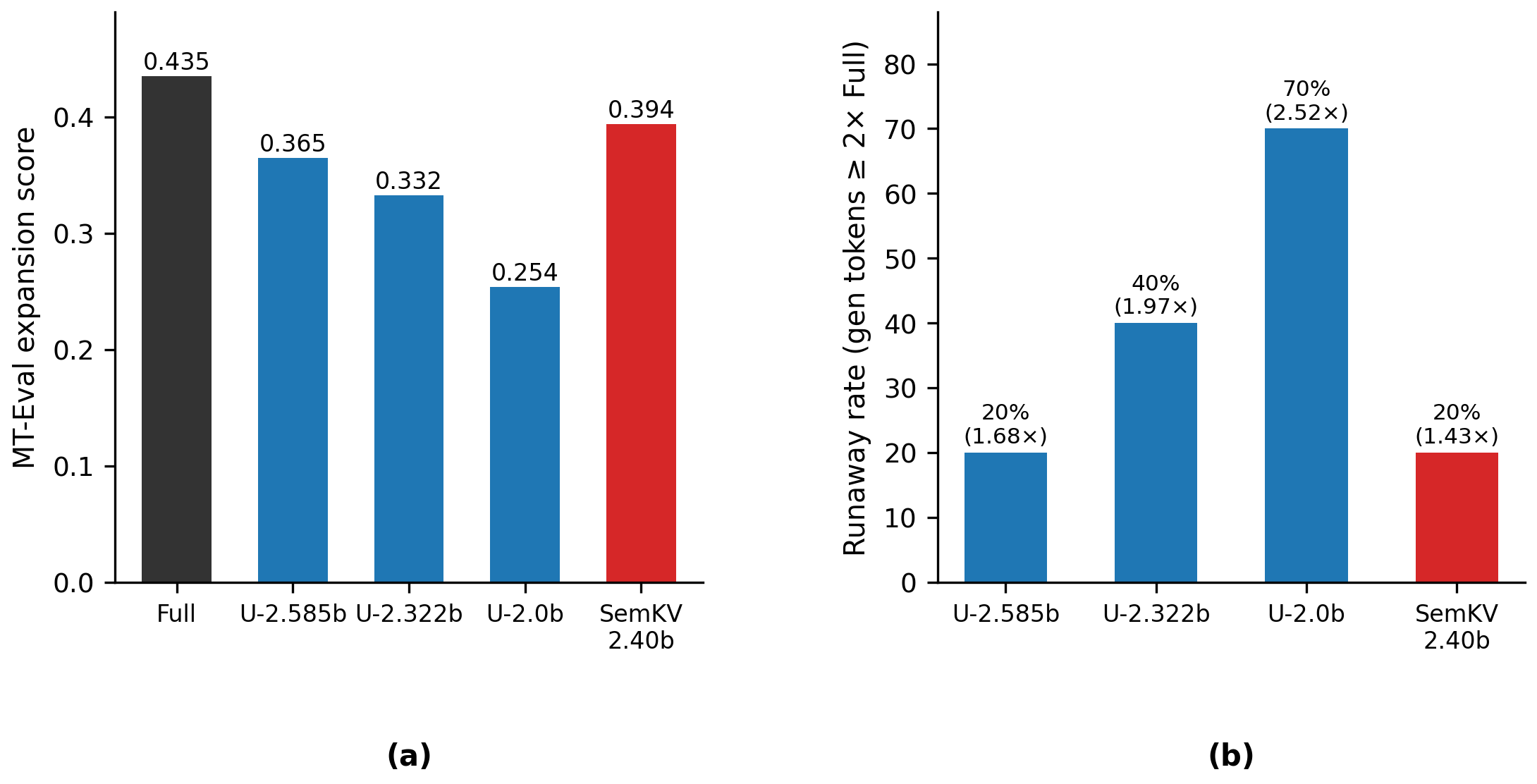}
\caption{Below-cliff stress (MT-Eval expansion, 10 conversations $\times$ 3 seeds). (a) At the operating point SemKV significantly outperforms both uniform anchors in this stress setting. (b) Runaway generation is suppressed by SemKV relative to the uniform anchors.}
\end{figure}

\begin{figure}[H]
\centering
\includegraphics[width=0.8\linewidth]{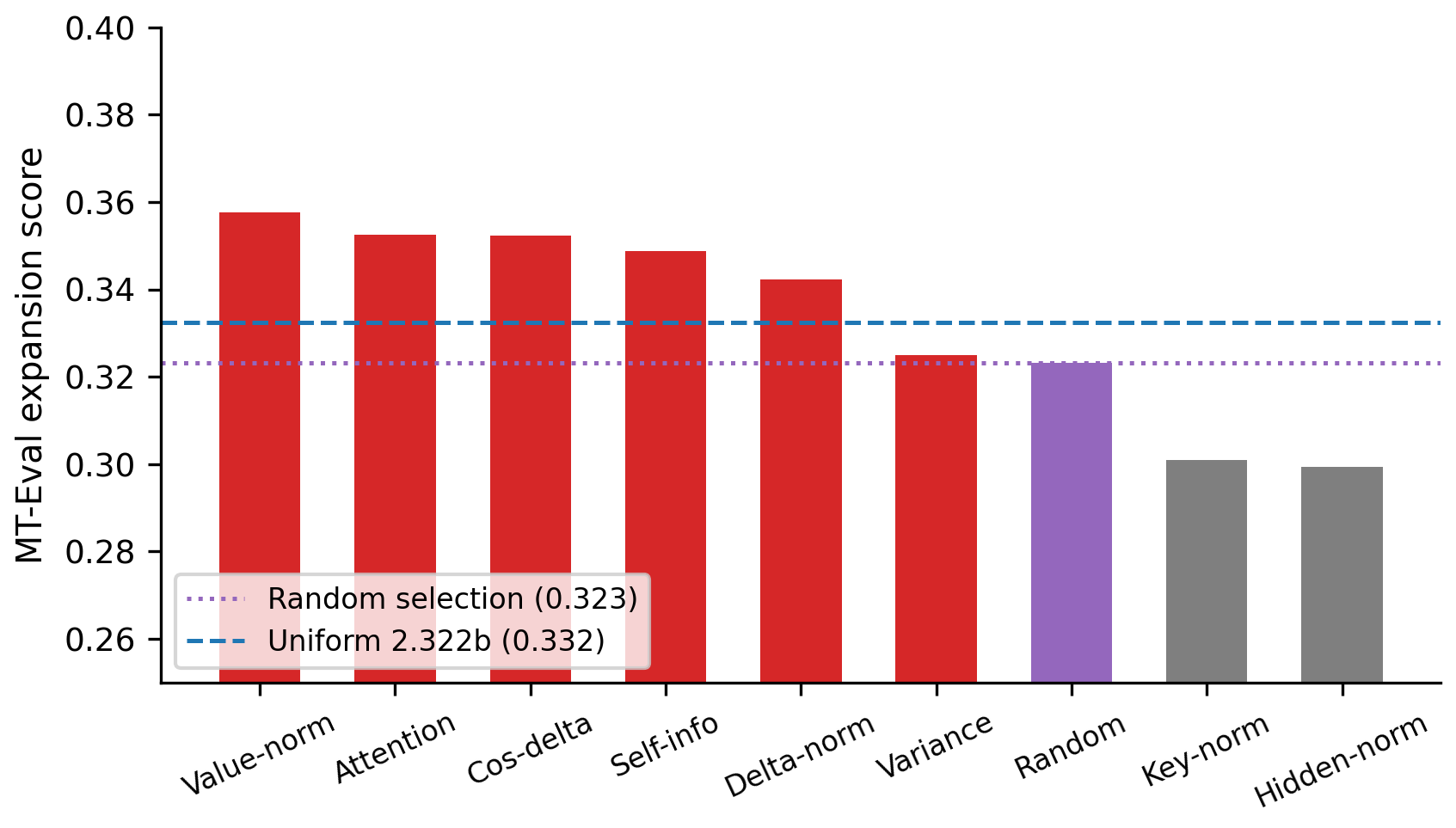}
\caption{Widened-gap indicator spectrum under stress (3.0/2.0 bits, average 2.30 bits, MT-Eval expansion). In this Llama stress setting, indicator quality separates when the low-bit side lies below the cliff: informative selection reduces catastrophic degradation when most tokens receive a below-cliff precision; indicators below the random line are worse than random.}
\end{figure}

\subsection{Cross-Model, Context-Length, and Base-Quantizer Transfer}
\label{sec:transfer}

Fig.~16 and Table~5 evaluate transfer to Mistral-7B-Instruct-v0.3 under the identical protocol (10 subsets $\times$ 30 $\times$ 3 seeds). Three properties transfer. (i) \emph{Cliff location}: only uniform 2.0-bit degrades significantly ($\Delta{=}-0.0249$, $p{<}10^{-3}$); 2.322--3.0 bits are indistinguishable from full KV. (ii) \emph{Grid interpolation}: SemKV at 2.39 bits shows no statistically detectable deficit from full KV for every informative indicator ($|\Delta|\le 0.012$, all n.s.\ after correction). (iii) \emph{Indicator indifference among informative indicators}: all pairwise comparisons between the eight model-internal indicators are n.s.\ (Bonferroni-corrected). Notably, on Mistral the mixing safety is \emph{not} indicator-free in the strong sense observed on Llama: uninformative (random) token selection at the same 2.39-bit budget degrades significantly ($\Delta{=}-0.0213$, $p{<}10^{-3}$, three-seed consistent), concentrated on retrieval-style tasks, while every informative indicator remains safe. Semantic selection is thus what makes aggressive mixing transferable across backbones. A three-seed context-budget sweep (2k/4k/7.5k tokens; Fig.~17) adds a further datapoint: at the 2k and 7.5k budgets only uniform 2.0-bit degrades, but at the mid budget (4k) uniform 2.322-bit also degrades significantly ($\Delta{=}-0.0243$, $p{<}10^{-4}$), and SemKV buffers this to roughly half the deficit ($\Delta{=}-0.0139$; SemKV vs.\ uniform 2.322-bit: $+0.0104$, $p{=}0.027$). SemKV is the setting closest to full KV at every budget. Absolute scores at smaller budgets include truncation effects and only within-budget differences are interpreted.

\begin{figure}[H]
\centering
\includegraphics[width=\textwidth]{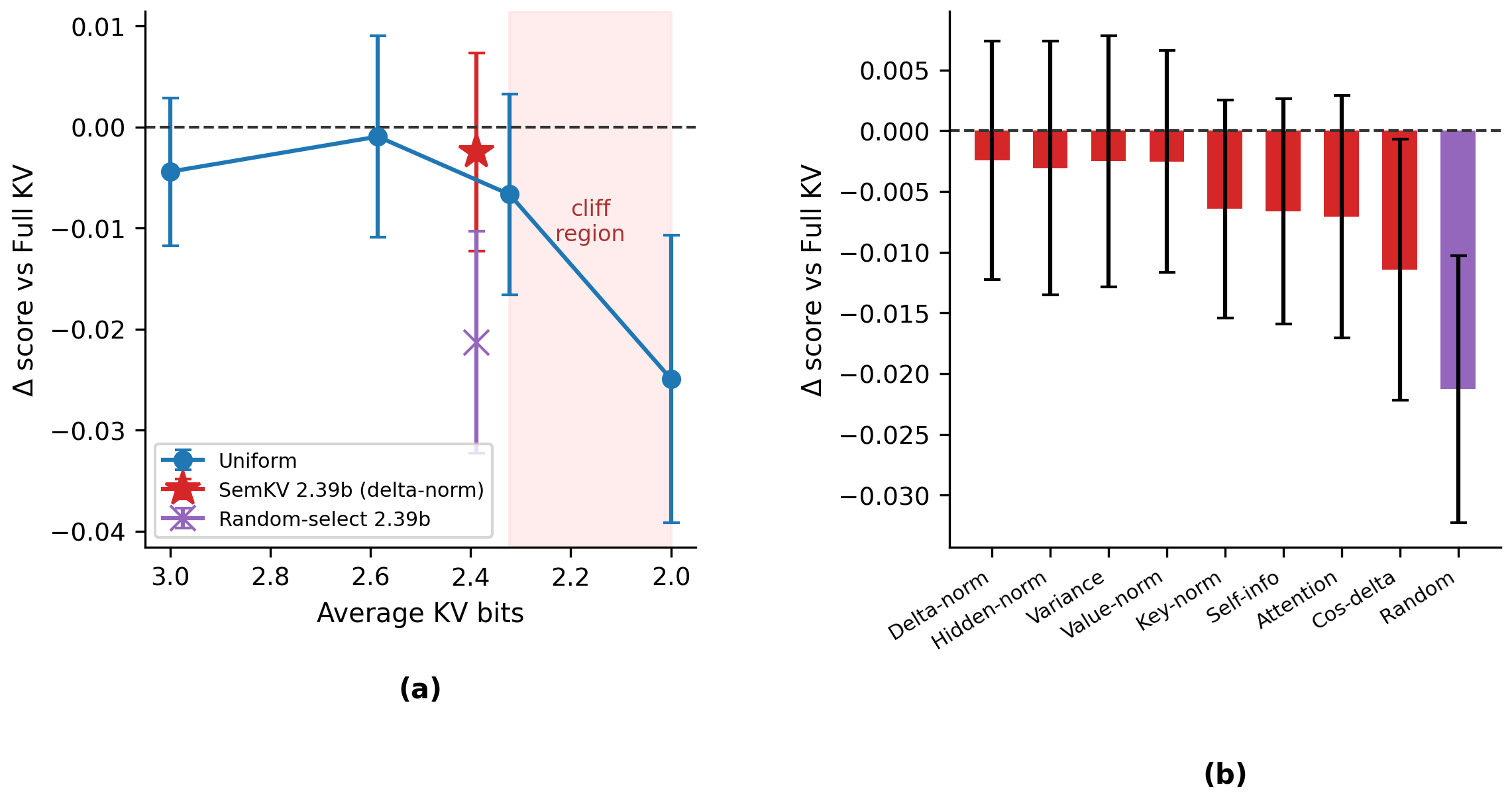}
\caption{Transfer to Mistral-7B-Instruct-v0.3 (10 subsets $\times$ 30 $\times$ 3 seeds). (a) The cliff transfers; SemKV at 2.39 bits remains statistically indistinguishable from full KV, whereas random selection at the same budget does not. (b) Per-indicator deltas vs.\ full KV. On Mistral, informative indicators remain mutually interchangeable, but random allocation is not safe --- indicator freedom is an operating-point-specific, not universal, result.}
\end{figure}

\begin{table}[H]
\centering
\caption{Cross-model transfer (paired $\Delta$ vs.\ full KV, $n{=}900$ per model). Absolute scores are not comparable across models. Compression is measured KV storage vs.\ FP16; the two backbones share the same KV shape, so the ratios apply to both.}
\small
\begin{tabular}{lccc}
\toprule
Method & Compr. & Llama-3.1-8B $\Delta$ & Mistral-7B $\Delta$ \\
\midrule
Uniform 3.0-bit & $4.9\times$ & $-0.0069$ (0.11) & $-0.0044$ (0.24) \\
Uniform 2.585-bit & $5.6\times$ & $-0.0094$ (0.09) & $-0.0009$ (0.86) \\
Uniform 2.322-bit & $6.2\times$ & $-0.0068$ (0.24) & $-0.0067$ (0.19) \\
Uniform 2.0-bit & $7.1\times$ & $-0.045$ ($p{<}10^{-6}$) & $-0.0249$ ($p{<}10^{-3}$) \\
SemKV 2.39-bit (delta-norm) & $6.0\times$ & $+0.0011$ (0.84) & $-0.0025$ (0.62) \\
\bottomrule
\end{tabular}
\end{table}

\begin{figure}[H]
\centering
\includegraphics[width=0.7\linewidth]{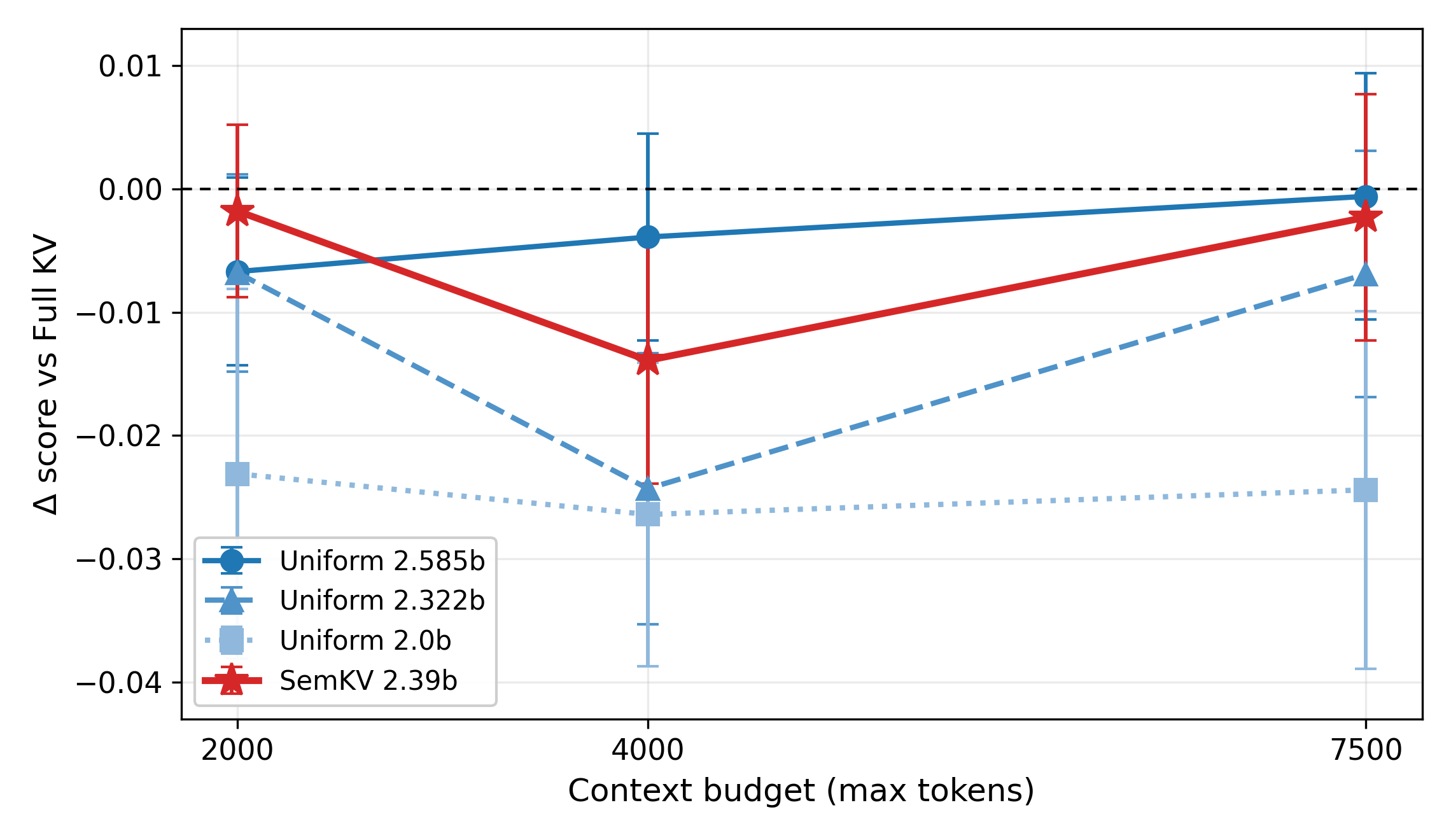}
\caption{Context-budget sweep on Mistral (3 seeds). Within each budget, SemKV stays closest to full KV among the quantized settings; absolute levels include truncation effects. The measured bracket can depend on context-budget composition even when the primary 7.5k protocol is stable.}
\end{figure}

\textbf{Base-quantizer transfer.} The cliff of Sec.~4.2 was mapped with a per-row asymmetric affine quantizer under Hadamard rotation. To test whether the cliff is a property of the model alone or of the \emph{model--quantizer pair}, we replace the base quantizer with a reimplementation of TurboQuant-MSE~[9] --- random rotation followed by a Gaussian-optimal Lloyd--Max codebook, extended here to non-power-of-two level counts so the fractional grid is preserved --- and repeat the uniform sweep and the SemKV operating point on the single-turn protocol (10 subsets $\times$ 30 $\times$ 3 seeds, shared full-KV baseline). Affine anchors run in the same sweep reproduce Sec.~4.2 exactly (uniform 2.322-bit n.s.; uniform 2.0-bit $\Delta{=}-0.0448$, $p{<}10^{-7}$), validating the port. Table~6 and Fig.~18 show the outcome: with the distortion-optimized base, \emph{the entire tested range $[2.0, 3.0]$ bits is flat}. Uniform 2.0-bit under TurboQuant is statistically indistinguishable from full KV ($\Delta{=}-0.0069$, $p{=}0.24$), a direct paired recovery of $+0.0379$ over affine 2.0-bit ($p{<}10^{-5}$), and the runaway signature of the cliff disappears with it: the max-length generation rate at 2.0 bits drops from 28.8\% under affine quantization to 19.9\% under TurboQuant (affine vs.\ TurboQuant: McNemar $p{<}10^{-11}$), statistically indistinguishable from the full-KV baseline of this run (21.8\%, McNemar $p{=}0.09$). Residual depth at 2.0 bits is seed-dependent (one seed shows $-0.027$; the other two are null), mirroring the depth seed-dependence noted in Sec.~4.2; a marginal uncorrected deficit at 3.0 bits ($p{=}0.02$) does not survive Bonferroni correction and is non-monotone in bits, so we treat it as noise. Two further observations complete the picture. First, \emph{orthogonality}: SemKV's token-axis interpolation on the TurboQuant base is statistically indistinguishable from full KV ($\Delta{=}-0.0065$, n.s.), is indistinguishable from uniform TurboQuant 2.322-bit, and closely matches the affine uniform 2.322-bit anchor ($\Delta{=}+0.0004$) --- the allocation mechanism composes with a better quantizer unchanged. Second, a same-protocol comparison with TurboQuant's own \emph{channel-axis} mixed-precision recipe (outlier channels at 3 bits, the rest at 2 bits; 2.5 code bits): quality is statistically indistinguishable from token-axis SemKV ($\Delta{=}-0.0042$, $p{=}0.48$), while SemKV's effective storage is smaller (2.525 vs.\ 2.750 bits/value, as TurboQuant-MSE stores one scalar per row against two for the channel-split configuration). The practical reading is not that either mixing axis dominates, but that the cliff location must be measured for the target deployment setting: with a stronger base quantizer the flat region extends downward --- here to at least 2.0 bits, i.e., a measured $7.5\times$ compression with no statistically detectable deficit on this protocol --- and SemKV's operating rule, interpolate above the measured cliff, carries over unchanged. Two follow-up sweeps complete the picture. \textbf{Locating the TurboQuant cliff.} Extending the uniform sweep below 2.0 bits (3-level and 2-level Lloyd--Max grids, same protocol and seeds) locates the cliff rather than merely bounding it: uniform 1.585-bit collapses ($\Delta{=}-0.0379$, $p{<}10^{-5}$, all three seeds consistent) and 1.0-bit is catastrophic ($-0.267$), so the TurboQuant cliff sits at $(1.585, 2.0]$ --- \emph{exactly one grid step below} the affine cliff of Sec.~4.2. The collapse decouples from the runaway signature: in the matched generation-side runs, max-length rates stay at the full-KV level of the sub-2-bit run (21.8\%) at both 1.585 bits (22.0\%, McNemar $p{=}0.92$) and 1.0 bit (21.4\%, McNemar $p{=}0.87$). The interior is shallower than the affine interior reported in Sec.~4.2: protecting 75\% of tokens at 2.0 bits (code 1.9 bits, effective 2.025, $7.9\times$) returns to statistical indistinguishability from full KV ($\Delta{=}-0.0091$, $p{=}0.17$), and mixing significantly outperforms the lower uniform grid point ($+0.018$ at $r{=}50\%$, $p{=}0.005$; $+0.029$ at $r{=}75\%$, $p{<}10^{-3}$), while delta-norm and random selection remain statistically indistinguishable inside the cliff ($p{=}0.072$) --- in contrast to the affine interior of Sec.~4.2, where delta-norm selection at $r{=}50\%$ was significantly \emph{worse} than random.

\textbf{Generation-time and multi-turn transfer.} Repeating the generation-quantization protocol of Sec.~4.4 and the multi-turn protocol of Sec.~4.6 with the TurboQuant base (3 seeds each; affine anchors reproduce Secs.~4.4 and 4.6 in the same runs: generation $-0.0157$, $p{<}10^{-5}$; multi-turn $-0.070$, $p{<}10^{-48}$) shows that the boundary shift is protocol-wide (Table~7, Fig.~19). On the generation side the cliff is fully bracketed: TurboQuant is safe at 2.322, 2.0, and 1.585 bits (all n.s.\ vs.\ FP16-generation; largest deficit $-0.0031$), and collapses at 1.0 bit ($\Delta{=}-0.0329$, $p{<}10^{-18}$, three seeds consistent) --- the generation-side cliff sits at $(1.0, 1.585]$, at least one grid step below its affine location, and the collapse again arrives without the runaway signature (max-length rates match FP16 generation, McNemar $p{=}0.87$). In multi-turn dialogue the collapse point moves exactly one grid step, mirroring prefill: uniform 1.585-bit collapses by $-0.0912$ ($p{<}10^{-79}$, deeper than the affine 2.0-bit collapse in the same run, paired $\Delta{=}-0.0212$), while 2.0-bit removes $68\%$ of the affine collapse ($-0.0222$ vs.\ $-0.070$).

Above the cliff, multi-turn retains the small residual floor already documented for the affine base in Sec.~4.6, and a dedicated same-run comparison shows it is \emph{not} specific to either quantizer: affine uniform 2.322-bit sits at $-0.0209$ ($p{<}10^{-10}$) and TurboQuant at $-0.0083$ ($p{<}10^{-3}$) in the same run --- the better base shrinks the floor by $2.5\times$ (paired $+0.0126$, $p{<}10^{-6}$) but does not remove it, and raising TurboQuant to 2.585 bits leaves the floor unchanged ($-0.0077$; vs.\ 2.322-bit $p{=}0.77$), so it is not fixable by spending more bits in this range. Selection behaves identically on both bases: the SemKV operating point is statistically indistinguishable from the upper uniform grid point ($p{=}0.68$ under TurboQuant) --- selection reaches the ceiling the protocol admits --- and above-cliff interpolation composes in multi-turn as well: mixing 2.322/2.0 at $r{=}35\%$ (token-weighted effective 2.236 bits/value, $7.16\times$) significantly improves on uniform 2.0-bit ($+0.0110$, $p{<}10^{-3}$) and is statistically indistinguishable from uniform 2.322-bit quality while using $0.10$ fewer code bits, exactly as the affine mix behaves relative to uniform 2.585-bit in Sec.~4.6. We do not treat the multi-turn mix as a no-detectable-loss operating point, because its residual deficit relative to full KV ($-0.0111$) remains significant under this protocol.

This decomposition also reconciles our sweep with the original TurboQuant report, which finds quality neutrality at 3.5 bits and \emph{marginal degradation already at 2.5 bits}: that claim concerns quantization of the entire cache including decode-side tokens, whereas the flat single-turn sweep above quantizes the prefill cache only. In our own full-cache protocol (multi-turn), both bases show exactly such small deficits above the cliff --- there is no contradiction, only a decomposition of where the loss lives. Across all three manifestations the conclusion is uniform: \emph{replacing the base quantizer moves the collapse boundary down by at least one grid step and leaves the operating rule --- map the cliff, interpolate above it --- unchanged.} Transfer to other backbones under the TurboQuant base is left untested (Sec.~5).

\begin{table}[H]
\centering
\caption{Base-quantizer transfer (Llama-3.1-8B, single-turn, $n{=}900$). Effective bits include packing and per-row metadata (Sec.~4.1: affine and channel-split store two scalars per row, TurboQuant-MSE one). Affine anchors reproduce Sec.~4.2 in the same run. The sub-2-bit rows (separate 3-seed run, shared protocol) locate the TurboQuant cliff at $(1.585, 2.0]$; the $r{=}75\%$ mix sits inside the affine-inaccessible gap between the 4-level and 3-level grids.}
\footnotesize
\setlength{\tabcolsep}{4pt}
\begin{tabular}{lcccc}
\toprule
Method & Code bits & Eff.\ bits & Score & $\Delta$ vs.\ full ($p$) \\
\midrule
Full KV & 16.0 & 16.0 & 0.4406 & --- \\
Affine uniform 2.322-bit & 2.333 & 2.583 & 0.4338 & $-0.0068$ (0.25) \\
Affine uniform 2.0-bit & 2.000 & 2.250 & 0.3958 & $-0.0448$ ($p{<}10^{-7}$) \\
TQ uniform 3.0-bit & 3.000 & 3.125 & 0.4296 & $-0.0110$ (0.02$^{\dagger}$) \\
TQ uniform 2.585-bit & 2.600 & 2.725 & 0.4347 & $-0.0059$ (0.23) \\
TQ uniform 2.322-bit & 2.333 & 2.458 & 0.4380 & $-0.0026$ (0.63) \\
TQ uniform 2.0-bit & 2.000 & 2.125 & 0.4337 & $-0.0069$ (0.24) \\
SemKV (TQ base) 2.39-bit & 2.400 & 2.525 & 0.4341 & $-0.0065$ (0.23) \\
TQ channel-mix 2.5-bit & 2.500 & 2.750 & 0.4383 & $-0.0023$ (0.60) \\
\midrule
TQ uniform 1.585-bit (3L) & 1.600 & 1.725 & 0.4027 & $-0.0379$ ($p{<}10^{-5}$) \\
TQ uniform 1.0-bit (2L) & 1.000 & 1.125 & 0.1739 & $-0.2667$ ($p{<}10^{-100}$) \\
SemKV (TQ) mix 2.0/1.585, $r{=}75\%$ & 1.900 & 2.025 & 0.4315 & $-0.0091$ (0.17) \\
\bottomrule
\end{tabular}

\smallskip
{\footnotesize $^{\dagger}$Does not survive Bonferroni correction ($\times 4 \Rightarrow p{=}0.08$) and is non-monotone in bits.}
\end{table}

\begin{table}[H]
\centering
\caption{Generation-time and multi-turn transfer of the TurboQuant base (3 seeds; generation: LongBench vs.\ FP16-generation baseline, $n{=}900$; multi-turn: MT-Eval five task types vs.\ full KV in a dedicated transfer run, $n{=}324$ paired conversation evaluations: 108 conversations $\times$ 3 seeds). Affine anchors reproduce Secs.~4.4/4.6 in the same runs. Multi-turn effective bits include per-row metadata; for mixed rows the reported rate uses the realized token-weighted high-precision fraction over the evaluated conversations rather than the nominal $r$ alone. Rows at generation 1.0-bit and multi-turn affine-2.322/TQ-2.585 come from a dedicated add-on run whose continuity anchors reproduce the main run exactly.}
\footnotesize
\setlength{\tabcolsep}{4pt}
\resizebox{\textwidth}{!}{%
\begin{tabular}{llccl}
\toprule
Protocol & Method & Eff.\ bits & $\Delta$ ($p$) & Note \\
\midrule
Gen.\ & Affine gen 2.0b & 2.250 & $-0.0157$ ($p{<}10^{-5}$) & anchor (Sec.~4.4) \\
Gen.\ & TQ gen 2.322b & 2.458 & $-0.0031$ (0.13) & safe \\
Gen.\ & TQ gen 2.0b & 2.125 & $+0.0004$ (0.88) & safe \\
Gen.\ & TQ gen 1.585b & 1.725 & $-0.0031$ (0.34) & safe \\
Gen.\ & TQ gen 1.0b & 1.125 & $-0.0329$ ($p{<}10^{-18}$) & gen cliff: $(1.0, 1.585]$ \\
\midrule
MT & Affine uniform 2.0b & 2.250 & $-0.0700$ ($p{<}10^{-48}$) & anchor (Sec.~4.6) \\
MT & Affine uniform 2.322b & 2.583 & $-0.0209$ ($p{<}10^{-10}$) & above-cliff floor (affine) \\
MT & TQ uniform 2.322b & 2.458 & $-0.0083$ ($p{<}10^{-3}$) & floor, $2.5\times$ smaller than affine \\
MT & TQ uniform 2.585b & 2.725 & $-0.0077$ ($p{<}10^{-3}$) & floor flat in bits ($p{=}0.77$) \\
MT & TQ uniform 2.0b & 2.125 & $-0.0222$ ($p{<}10^{-11}$) & $68\%$ of collapse removed \\
MT & TQ uniform 1.585b & 1.725 & $-0.0912$ ($p{<}10^{-79}$) & cliff (one step below affine) \\
MT & SemKV (TQ) 2.585/2.322 r35 & 2.537 & $-0.0074$ ($p{<}10^{-3}$) & $=$ u2.322 ceiling ($p{=}0.68$) \\
MT & SemKV (TQ) 2.322/2.0 r35 & 2.236 & $-0.0111$ ($p{<}10^{-5}$) & $+0.011$ vs.\ u2.0 ($p{<}10^{-3}$) \\
\bottomrule
\end{tabular}}%
\end{table}

\begin{figure}[H]
\centering
\includegraphics[width=0.9\textwidth]{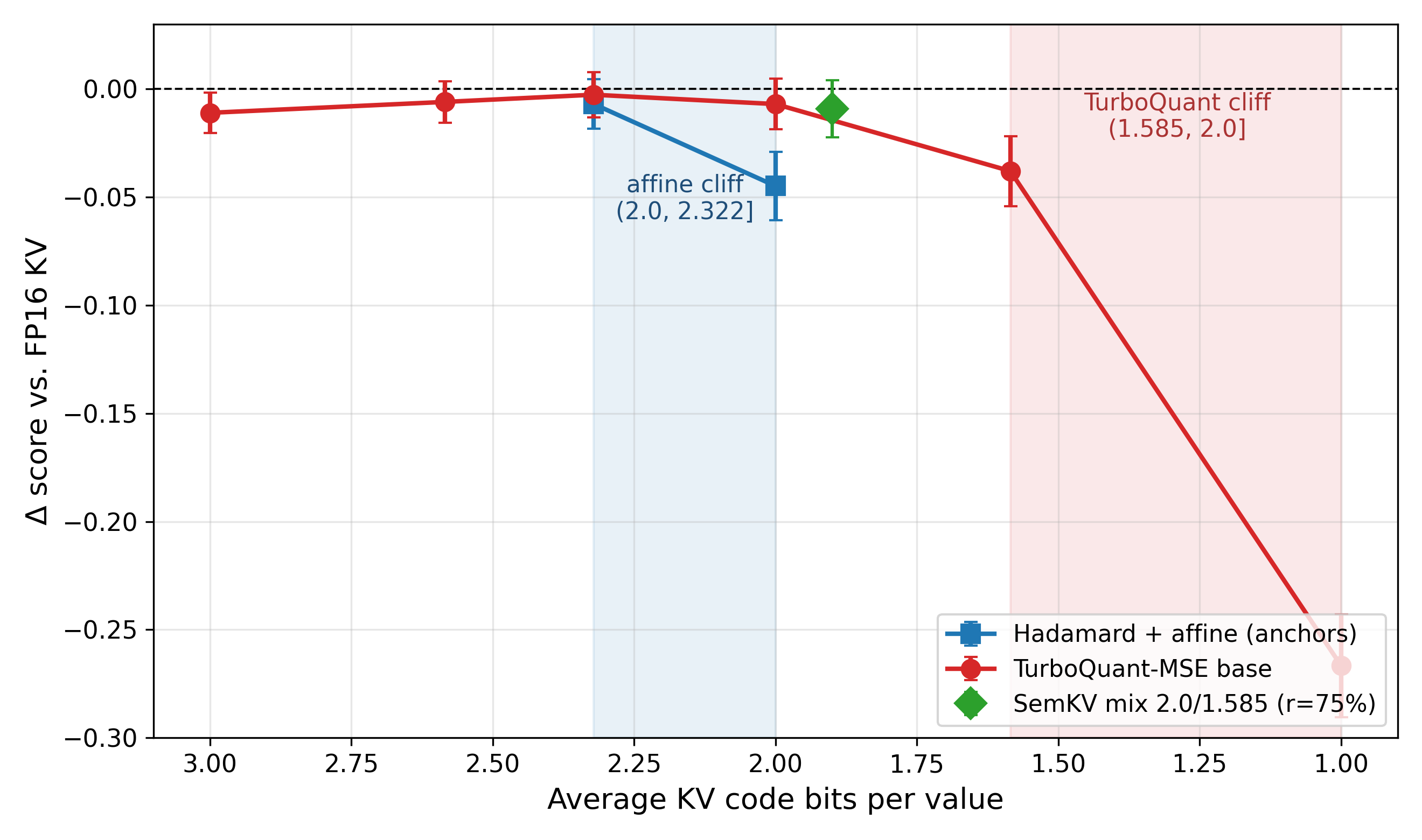}
\caption{Base-quantizer transfer: single-turn uniform sweep from 3.0 down to 1.0 code bits (3 seeds; error bars: 95\% CI of the paired difference). The affine cliff at $(2.0, 2.322]$ moves one grid step down to $(1.585, 2.0]$ under TurboQuant-MSE; the SemKV $r{=}75\%$ mix occupies the affine-inaccessible gap at code 1.9 bits, statistically indistinguishable from full KV.}
\end{figure}

\begin{figure}[H]
\centering
\includegraphics[width=0.9\textwidth]{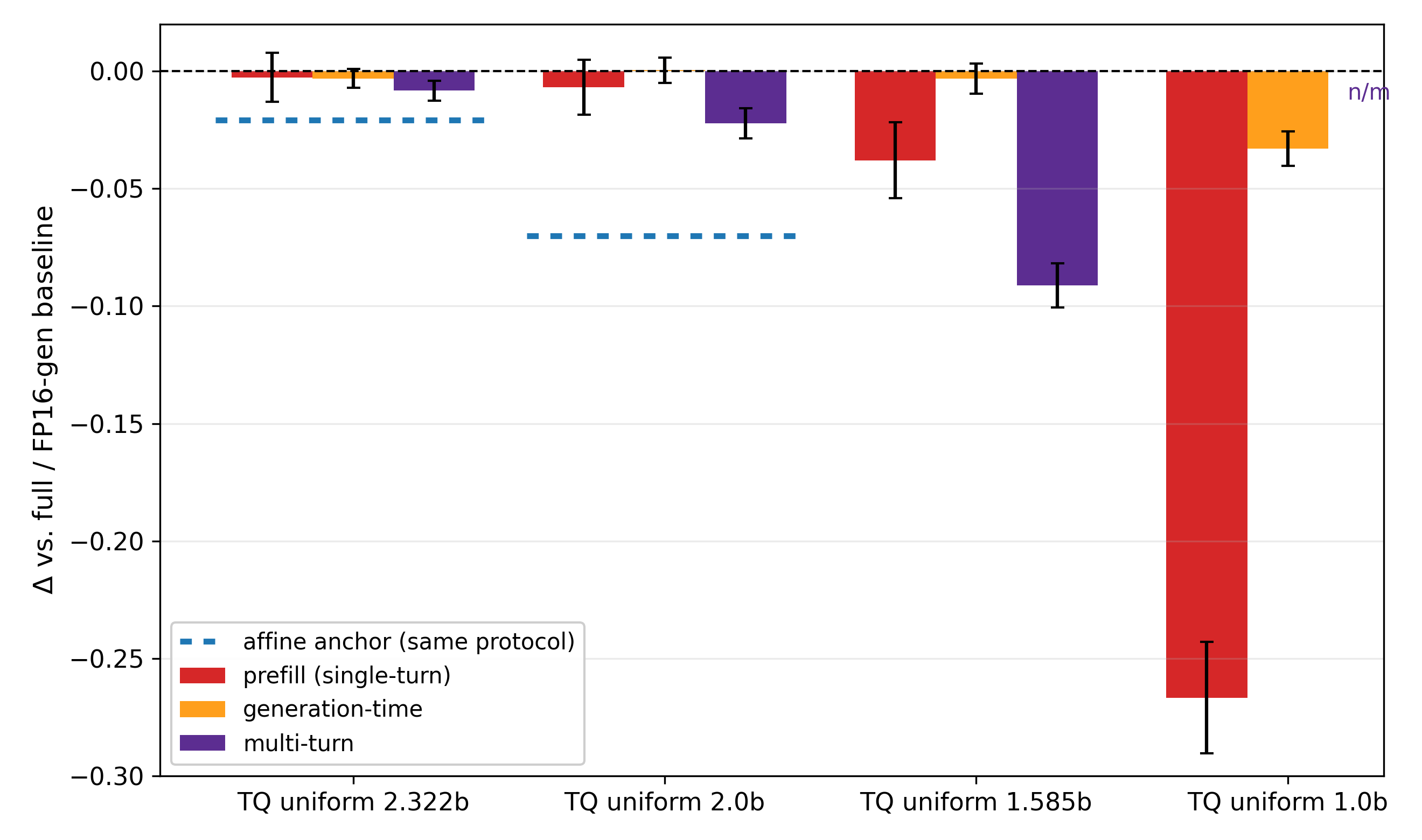}
\caption{Base-quantizer transfer across protocols (3 seeds; error bars: 95\% CI). The downward shift transfers across all three protocols: TurboQuant collapses at 1.585 bits in prefill and multi-turn and at 1.0 bit in generation (n/m: multi-turn not measured at 1.0 bit). Dotted blue markers give same-protocol affine anchors --- the 2.0-bit collapse anchors of each protocol, and the multi-turn affine 2.322-bit floor, which TurboQuant shrinks $2.5\times$.}
\end{figure}

\FloatBarrier

\section{Discussion}

\textbf{What mixed precision is for.} The cliff map resolves an apparent tension in the literature: importance-aware allocation sometimes helps greatly and sometimes not at all. In the flat regime above the cliff, preserving all tokens at above-cliff precisions is sufficient and the tested indicators are empirically interchangeable (a Llama operating-point result; Sec.~4.9 shows that random allocation is not safe on Mistral) --- there, the value of SemKV is \emph{grid interpolation}, reaching average precisions uniform quantization cannot realize. At and below the cliff, structure alone no longer protects quality and selection becomes decisive, to the point that a poorly aligned indicator is worse than random. The interior probes make the below-cliff half of this picture concrete: recovery is monotone but slow in the protected fraction, and concentrating protection on high-importance tokens is worse than spreading it, because below-cliff degradation is broad-based rather than evidence-localized. This two-regime view predicts, and our interior, stress, and transfer results confirm, that the robust operating recipe is: map the collapse cliff for the target \emph{deployment setting} and interpolate on its high-precision side; whether the resulting point is statistically ``safe'' must still be evaluated under the target protocol. If an aggressive configuration includes a below-cliff component, informative selection becomes important for limiting degradation. The base-quantizer transfer of Sec.~4.9 makes the pair-dependence concrete: a distortion-optimized base (TurboQuant-MSE) moves the collapse boundary down by one grid step in prefill and multi-turn --- and by at least that much on the generation side --- while the interpolation mechanism and its quality carry over unchanged; the one qualification is multi-turn, where a small above-cliff floor persists under \emph{either} base --- the affine floor is $2.5\times$ deeper than TurboQuant's, more bits do not remove it, and its task profile mirrors the collapse ordering --- setting a ceiling that selection reaches (remaining statistically indistinguishable from the upper grid point while using fewer bits on both bases) but cannot exceed. This floor is also what reconciles the flat prefill-only sweep with the original TurboQuant report of marginal degradation at 2.5 bits on the full cache. The cliff map is thus not a fixed constant of the model but a measurement the recipe prescribes; better quantizers lower the cliff and thereby raise the compression available above it.

\textbf{Deletion versus precision redistribution.} SemKV redistributes precision instead of deleting tokens, preserving the attention graph. The controlled comparison shows that deletion fails even with a larger memory budget and across the tested selection indicators; low-precision preservation of every token is the operative mechanism, consistent with observations that retaining evicted pairs at low precision recovers most eviction damage~[71].

\textbf{Limitations.} Under the affine base, the primary 7.5k evaluations show a consistent catastrophic transition at 2.0 bits across the tested protocols and two backbones; however, the measured bracket can vary with context-budget composition (the 4k setting in Sec.~4.9), evaluation protocol (the generation-side TurboQuant boundary sits one step lower), and base quantizer, and the cliff \emph{depth} is seed- and task-dependent. Moreover, the stress evidence for indicator separation rests on a small below-cliff sample. The comparison to pruning is a structural control under a shared indicator, not a benchmark against engineered eviction systems with recency windows and sink handling. Scoring-cost and storage measurements are microbenchmarks on a single GPU class, and the storage figures assume the mixed-radix packed layout of Sec.~4.1 rather than an optimized fused kernel. The base-quantizer transfer (Sec.~4.9) now covers prefill, generation-time, and multi-turn protocols on Llama, but not other backbones; the TurboQuant comparison uses our reimplementation with documented choices where the original leaves details unspecified (outlier-channel selection; rotation sharing). Full-cache multi-turn quantization retains a small above-cliff floor under both bases that no allocation policy we tested removes; the compounding-through-dialogue interpretation of this floor is inferred from its task profile and bits-insensitivity rather than directly ablated; and the 2.322/2.0 multi-turn interpolation point, while superior to uniform 2.0-bit, is not treated as a no-detectable-loss operating point. Extending the cliff map to more backbones and to KV-reuse serving stacks is left to future work.

\FloatBarrier

\section{Conclusion}

This paper proposed SemKV, an all-token-preserving mixed-precision KV cache compression framework grounded in an empirical quality-cliff map of uniform KV quantization on a fractional-bit grid. On Llama-3.1-8B under the affine base, the cliff lies in $(2.0, 2.322]$ bits and co-occurs with runaway generation; SemKV interpolates strictly above it and shows no statistically detectable deficit from full KV at 2.39 average bits on our LongBench protocol --- a measured $6.0\times$ KV-cache storage reduction versus FP16, metadata included --- for every one of eight model-internal importance indicators. The block-exact extension carries the result to generation-time tokens and, up to a small above-cliff multi-turn floor that persists under both quantizers (with magnitude $2.5\times$ smaller under TurboQuant-MSE), to multi-turn dialogue; a controlled comparison shows deletion-based compression failing at a $1.5\times$ larger measured memory budget where low-precision preservation succeeds; a below-cliff stress test shows where indicator quality begins to matter; and the cliff, the interpolation property, and the statistically indistinguishable mixed-precision operating point transfer to Mistral-7B, where informative selection is what keeps aggressive mixing safe. The cliff, finally, is an empirical property of the target deployment setting rather than a constant of the model: swapping the affine base for TurboQuant-MSE lowers the collapse boundary by at least one grid step in prefill, generation, and multi-turn alike, and SemKV converts the extra headroom directly into compression --- interpolating at 2.0/1.585 bits inside the grid gap that the affine base cannot even reach, it is statistically indistinguishable from full KV at an effective 2.025 bits/value, improving the no-detectable-loss operating point from $6.0\times$ to $7.9\times$. These results suggest that better base quantizers are inputs to SemKV rather than competitors: when a stronger base lowers the measured cliff, the same rule --- map the cliff, interpolate above it --- converts that shift into additional memory savings.


\section*{Declaration of competing interest}
The authors declare the following financial interests/personal relationships which may be considered as potential competing interests: A patent application related to the method described in this manuscript has been filed by ETRI (Korean patent application, 2026).

\section*{Declaration of Generative AI and AI-assisted technologies in the writing process}
During the preparation of this work, the author used generative AI and AI-assisted technologies solely to assist with English grammar, wording, and readability. The author reviewed and edited all AI-assisted suggestions and takes full responsibility for the content of the publication.

\section*{Data availability}
The benchmark datasets used in this study (LongBench, MT-Eval) are publicly available. Code and experiment configurations will be made available upon reasonable request.

\section*{Acknowledgments}
This work was supported by the Institute of Information \& Communications Technology Planning \& Evaluation (IITP) grant funded by the Korea government (MSIT) (RS-2024-00336738, Development of Complex Task Planning Technologies for Autonomous Agents).

\end{document}